\RequirePackage{fix-cm}
\documentclass[twocolumn]{svjour3}           
\smartqed  
\usepackage{graphicx}
\usepackage{graphicx}
\usepackage{booktabs}
\usepackage{bm}
\usepackage{diagbox}
\usepackage{latexsym}
\usepackage[misc]{ifsym}
\usepackage{amsmath}
\usepackage{amssymb}
\usepackage{subfig}
\usepackage[colorlinks,linkcolor=blue,anchorcolor=blue,citecolor=blue, urlcolor=blue]{hyperref}
\begin{document}

\title{Rotation Differential Invariants of Images Generated by Two Fundamental Differential Operators
}


\author{Hanlin Mo\textsuperscript{1,2}\and
        Hua Li\textsuperscript{1,2}
}

\authorrunning{Hanlin Mo et al.} 

\institute{Hanlin Mo\textsuperscript{~\Letter} \at
              \email{mohanlin@ict.ac.cn}
           \and
           Hua Li \at
             \email{lihua@ict.ac.cn}
           \\\\\textsuperscript{1} Key Laboratory of Intelligent Information Processing, Institute of Computing Technology, Chinese Academy of Sciences, Beijing, China\\\\
           \textsuperscript{2} University of Chinese Academy of Sciences, Beijing, China
}

\date{Received: date / Accepted: date}

\maketitle

\begin{abstract}
In this paper, we design two fundamental differential operators for the derivation of rotation differential invariants of images. Each differential invariant obtained by using the new method can be expressed as a homogeneous polynomial of image partial derivatives, which preserve their values when the image is rotated by arbitrary angles. We produce all possible instances of homogeneous invariants up to the given order and degree, and discuss the independence of them in detail. As far as we know, no previous papers have published so many explicit forms of high-order rotation differential invariants of images. In the experimental part, texture classification and image patch verification are carried out on popular real databases. These rotation differential invariants are used as image feature vector. We mainly evaluate the effects of various factors on the performance of them. The experimental results also validate that they have better performance than some commonly used image features in some cases.

\keywords{Rotation differential invariants \and Homogeneous polynomials \and Independent sets \and Patch verification \and Texture classification}
\end{abstract}

\section{Introduction}
\label{sec:1}
How to extract effective features of images is one of the most fundamental problems in the fields of computer vision and pattern recognition. A desirable feature should be able to capture intrinsic information of objects of interest in images. This implies that it must be naturally invariant to various kinds of geometric deformations caused by imaging geometry. Two-dimensional rotation is the most common image deformation. For decades, researches have designed numerous rotation invariant features of images for different applications. Among them, rotation differential invariants play a very important role.

Given an image function, its rotation differential invariants are a series of functions of its partial derivatives. The order of an invariant is the maximal order of derivatives it depends upon. For example, the Gaussian and the mean curvature at a point of an image are two rotation differential invariants of order $2$. Koenderink and Van Doom used them to define two measures of image local shape, the 'curvedness' and the 'shape index'\cite{12}. The Hessian matrix is built with second-order partial derivatives. Its determinant and its trace are also second-order rotation differential invariants. Lindeberg et~al. and Mikolajczyk et~al. found that the former is suitable to detect blob structures in images, and the latter has the ability to determine the characteristic scale of each structure \cite{13,14,15,16,17,18}. Griffin et~al. made use of second-order rotation differential invariants to approximate symmetry types of image local regions, and to construct the basic image feature for texture classification \cite{9,10,11}.

Compared to second-order rotation differential invariants, few studies focused on higher-order rotation differential invariants of images. In fact, it is not very easy to systematically generate them and to assign proper geometrical meanings to them. Olver did outstanding work in this filed. He published numerous books and papers to explain how to use the equivariant method of moving frames to produce higher-order rotation differential invariants of images \cite{20,22}. Florack et~al. also derived higher-order rotation differential invariants by repeatedly applying two directional differential operators to an image function \cite{7,8,24}. Note that most of invariants generated by using these methods depend upon the first order derivatives \cite{24}.

In this paper, we propose a new method to algorithmically derive higher-order rotation differential invariants of images by means of two fundamental differential operators. Each of invariants obtained by using our method can be expressed as a homogeneous polynomial of image partial derivatives, and the degree of its expression is called its degree. In theory, rotation differential invariants can be arbitrary types of functions of image partial derivatives, such as polynomials, rational polynomials and so on. However, Florack et~al. have proved that any non-polynomial rotation differential invariants can be expressed in terms of homogeneous polynomial ones \cite{7,8}. Thus, compared to previous invariants, homogeneous invariants produced in our paper are more fundamental. They can be called primitive invariants. We generate all possible homogeneous invariants up to the degree $4$ and the order $4$. And then, linear dependencies, polynomial dependencies and functional dependencies among them are identified, and we thus derived three types of independent sets. The patch verification and texture classification are conducted on popular image databases. We analysed the influence of various factors on the performance of our rotation differential invariants, and compared them with some widely used image features.

The paper is organized as follow. Section \ref{sec:2} provides needed definitions and notations for our work. Section \ref{sec:3} and \ref{sec:4} are the main contribution of this paper. We design two fundamental differential operators and use them to systematically generate rotation differential invariants of images. In Section \ref{sec:5}, we give some invariants in explicit forms and study mutual independence of them. In order to better validate our work, the corresponding experiments are carried out in Section \ref{sec:6}. Section \ref{sec:7} concludes our work and discuss potential future work.

\section{Basic Definitions and Notations}
\label{sec:2}
In this section, we introduce basic definitions and notations often used in the following sections.
\subsection{Rotation of Images}
\label{sec:2.1}
An image can be defined as a 2D scalar function
\begin{equation}\label{equ:1}
f(x,y):\Omega \subset \mathbb{R}^{2} \rightarrow \mathbb{R}
\end{equation}
Suppose that $h(u,v): \Omega^{'}\subset\mathbb{R}^{2}\rightarrow \mathbb{R}$ is a rotated version of $f(x,y)$, we have $h(u,v)=f(x,y)$ where
\begin{equation}\label{equ:2}
\left(
  \begin{array}{c}
    u \\
    v \\
  \end{array}
\right)
=R_{\theta}
\left(
  \begin{array}{c}
    x \\
    y \\
  \end{array}
\right)
=
\left(
  \begin{array}{cc}
    \cos\theta & -\sin\theta \\
    \sin\theta & \cos\theta \\
  \end{array}
\right)
\left(
  \begin{array}{c}
    x \\
    y \\
  \end{array}
\right)
\end{equation}
Since the rotation matrix $R_{\theta}$ has the properties ${R_{\theta}}^{-1}={R_{\theta}}^{T}$ and $det(R_{\theta})=1$, we can directly derive the inverse transformation of (\ref{equ:1})
\begin{equation}\label{equ:3}
\left(
  \begin{array}{c}
    x \\
    y \\
  \end{array}
\right)
={R_{\theta}}^{T}
\left(
  \begin{array}{c}
    u \\
    v \\
  \end{array}
\right)
=
\left(
  \begin{array}{cc}
    \cos\theta & \sin\theta \\
    -\sin\theta & \cos\theta \\
  \end{array}
\right)
\left(
  \begin{array}{c}
    u \\
    v \\
  \end{array}
\right)
\end{equation}

\subsection{Rotation Differential Invariants of Images}
\label{sec:2.2}
Given an image $f(x,y)$, its rotation differential invariant $RI$ can be expressed as the function of its partial derivatives $f_{pq}=\frac{\partial^{p+q}{f}}{\partial{x^{p}}\partial{y^{q}}}$, where $p$ and $q$ are non-negative integers. For any rotated versions $h(u,v)$ of $f(x,y)$, we have $RI^{'}=RI$, where $RI^{'}$ is obtained by replacing $f_{pq}$ in $RI$ with $h_{pq}=\frac{\partial^{p+q}{h}}{\partial{u^{p}}\partial{v^{q}}}$.

As mentioned previously, we focus on generating $RI$ which takes the form of homogeneous polynomial in $f_{pq}$, meaning that
\begin{equation}\label{equ:4}
RI=\sum^{M}_{m=1}c_{m}\prod^{N}_{n=1}f_{p_{mn}q_{mn}}
\end{equation}
where the positive integer $M$ denotes the number of monomials, all coefficient $c_{m}$ are non-zero integers, and $p_{mn}$ and $q_{mn}$ are non-negative integers. As stated previously, the degree of $RI$ is $N$, and the order of $RI$ is $K=\max\limits_{m,n}\{p_{mn}+q_{mn}\}$.

\section{Two Fundamental Differential Operators}
\label{sec:3}
In this section, we define two fundamental differential operators which have invariance to rotation of 2D spatial coordinates. They can be advantageously employed to derive rotation differential invariants of images, as shown in the next section.
\\
\\
\noindent\textbf{Definition 1.~} Let the symbols $\nabla_{i}$ and $\nabla_{j}$ denote the gradient operators with respect to $(x_{i},y_{i})$ and $(x_{j},y_{j})$, respectively.
\begin{equation}\label{equ:5}
\begin{split}
&\nabla_{i}=
\left(
  \begin{array}{c}
    \frac{\partial}{\partial{x_{i}}} \\
    \frac{\partial}{\partial{y_{i}}} \\
  \end{array}
\right)
\\&
\nabla_{j}=
\left(
  \begin{array}{c}
    \frac{\partial}{\partial{x_{j}}} \\
    \frac{\partial}{\partial{y_{j}}} \\
  \end{array}
\right)
\end{split}
\end{equation}
Then, two fundamental differential operators $\phi_{ij}$ and $\psi_{ij}$ can be defined by
\begin{equation}\label{equ:6}
\begin{split}
\phi_{ij}&={\nabla_{i}}^{T}\cdot \nabla_{j}\\&
=\left(\frac{\partial}{\partial{x_{i}}},\frac{\partial}{\partial{y_{i}}}\right)\cdot
\left(
  \begin{array}{c}
    \frac{\partial}{\partial{x_{j}}} \\
    \frac{\partial}{\partial{y_{j}}} \\
  \end{array}
\right)\\&
=\frac{\partial^{2}}{\partial{x_{i}}\partial{x_{j}}}+\frac{\partial^{2}}{\partial{y_{i}}\partial{y_{j}}}\\
\end{split}
\end{equation}
and
\begin{equation}\label{equ:7}
\begin{split}
\psi_{ij}&=det\left(
\left(
 \begin{array}{cc}
    \nabla_{i} & \nabla_{j} \\
  \end{array}
\right)
\right)\\&
=
\left|
  \begin{array}{cc}
    \frac{\partial}{\partial{x_{i}}} & \frac{\partial}{\partial{x_{j}}} \\
     \frac{\partial}{\partial{y_{i}}} & \frac{\partial}{\partial{y_{j}}} \\
  \end{array}
\right|\\&
=\frac{\partial^{2}}{\partial{x_{i}}\partial{y_{j}}}-\frac{\partial^{2}}{\partial{x_{j}}\partial{y_{i}}}\\
\end{split}
\end{equation}
In particular, $\phi_{ij}=\phi_{ji}$, $\psi_{ij}=-\psi_{ji}$ and $\psi_{ii}\equiv0$.
\\
\\
\noindent\textbf{Theorem 1.~} Suppose $(x_{i},y_{i})$ and $(x_{j},y_{j})$ are transformed using (\ref{equ:1}) into $(u_{i},v_{i})$ and $(u_{j},v_{j})$, respectively. Let $\phi^{'}_{ij}$ and $\psi^{'}_{ij}$ be defined as
\begin{equation}\label{equ:8}
\begin{split}
&\phi^{'}_{ij}={\nabla^{'}_{i}}^{T}\cdot {\nabla^{'}_{j}}\\&
\psi^{'}_{ij}=det\left(\left(
  \begin{array}{cc}
    {\nabla^{'}_{i}} & {\nabla^{'}_{j}} \\
  \end{array}
\right)\right)\\
\end{split}
\end{equation}
where
\begin{equation}\label{equ:9}
\begin{split}
&\nabla^{'}_{i}=
\left(
  \begin{array}{c}
    \frac{\partial}{\partial{u_{i}}} \\
    \frac{\partial}{\partial{v_{i}}} \\
  \end{array}
\right)
\\&
\nabla^{'}_{j}=
\left(
  \begin{array}{c}
    \frac{\partial}{\partial{u_{j}}} \\
    \frac{\partial}{\partial{v_{j}}} \\
  \end{array}
\right)
\end{split}
\end{equation}
Then, we have $\phi^{'}_{ij}=\phi_{ij}$ and $\psi^{'}_{ij}=\psi_{ij}$, where $\phi_{ij}$ and $\psi_{ij}$ are defined by (\ref{equ:6}) and (\ref{equ:7}), respectively.
\\
\\
\noindent\textbf{Proof.~} In fact, according to (\ref{equ:9}) and the chain's rule for the compound function, the gradient operators $\nabla^{'}_{i}$ and $\nabla_{i}$ are related by
\begin{equation}\label{equ:10}
\begin{split}
\nabla^{'}_{i}&=\left(\frac{\partial}{\partial{u_{i}}},\frac{\partial}{\partial{v_{i}}}\right)^{T}\\&
=\left(\frac{\partial}{\partial{x_{i}}}\frac{\partial{x_{i}}}{\partial{u_{i}}}
+\frac{\partial}{\partial{y_{i}}}\frac{\partial{y_{i}}}{\partial{u_{i}}},
\frac{\partial}{\partial{x_{i}}}\frac{\partial{x_{i}}}{\partial{v_{i}}}
+\frac{\partial}{\partial{y_{i}}}\frac{\partial{y_{i}}}{\partial{v_{i}}}\right)^{T}\\&
=
\left(
  \begin{array}{cc}
    \frac{\partial{x_{i}}}{\partial{u_{i}}} & \frac{\partial{y_{i}}}{\partial{u_{i}}} \\
    \frac{\partial{x_{i}}}{\partial{v_{i}}} & \frac{\partial{y_{i}}}{\partial{v_{i}}}\\
  \end{array}
\right)
\left(
  \begin{array}{c}
    \frac{\partial}{\partial{x_{i}}} \\
    \frac{\partial}{\partial{y_{i}}} \\
  \end{array}
\right)
\\&
=
\left(
  \begin{array}{cc}
    \cos \theta & \sin \theta \\
    -\sin \theta & \cos \theta\\
  \end{array}
\right)
\left(
  \begin{array}{c}
    \frac{\partial}{\partial{x_{i}}} \\
    \frac{\partial}{\partial{y_{i}}} \\
  \end{array}
\right)
\\&
={R_{\theta}}^{T}\nabla_{i}\\
\end{split}
\end{equation}
Substituting the relations $\nabla^{'}_{i}={R_{\theta}}^{T}\nabla_{i}$ and $\nabla^{'}_{j}={R_{\theta}}^{T}\nabla_{j}$ into (\ref{equ:8}), we get:
\begin{equation}\label{equ:11}
\begin{split}
\phi^{'}_{ij}&={\nabla^{'}_{i}}^{T}\cdot{\nabla^{'}_{j}}\\&
=\left({R_{\theta}}^{T}\nabla_{i}\right)^{T}{R_{\theta}}^{T}\nabla_{j}\\&
={\nabla_{i}}^{T}R_{\theta}{R_{\theta}}^{T}\nabla_{j}\\&
={\nabla_{i}}^{T}R_{\theta}{R_{\theta}}^{-1}\nabla_{j}\\&
={\nabla_{i}}^{T}\nabla_{j}\\&
=\phi_{ij}\\
\end{split}
\end{equation}
and
\begin{equation}\label{equ:12}
\begin{split}
\psi^{'}_{ij}&=det\left(\left(
  \begin{array}{cc}
    {\nabla^{'}_{i}} & {\nabla^{'}_{j}} \\
  \end{array}
\right)\right)\\&
=det\left(\left({R_{\theta}}^{T}\nabla_{i},{R_{\theta}}^{T}\nabla_{j}\right)\right)\\&
=det\left({R_{\theta}}^{T}\left(\nabla_{i},\nabla_{j}\right)\right)\\&
=det({R_{\theta}}^{T})\cdot det\left(\left(\nabla_{i},\nabla_{j}\right)\right)\\&
=\psi_{ij}\\
\end{split}
\end{equation}
The proof is completed. \qed

Theorem 1 indicates that two fundamental differential operators $\phi_{ij}$ and $\psi_{ij}$ are invariant to rotation of 2D spatial coordinates.

\section{The Construction of Rotation Differential Invariants}
\label{sec:4}
Using two fundamental differential operators defined in Section \ref{sec:3}, we design a new method to generate rotational differential invariants of images.
\\
\\
\noindent\textbf{Definition 2.~} Given a positive integer $N$, we can define the cumulative product of two fundamental differential operators with respect to $(x_{n},y_{n})$, where $n=1,2,...,N$,
\begin{equation}\label{equ:13}
\begin{split}
D=\prod_{1\leq i< j\leq N}(\psi_{ij})^{t_{ij}}
\prod_{1\leq i\leq j\leq N}(\phi_{ij})^{s_{ij}}
\end{split}
\end{equation}
Note that $s_{ij}$ and $t_{ij}$ are non-negative integers, and each of $(x_{n},y_{n})$ must be involved at least once in the differential operator $D$.

According to (\ref{equ:6}) and (\ref{equ:7}), (\ref{equ:13}) can be expressed as a homogeneous polynomial of $N$ degree in partial derivative operators to $(x_{n},y_{n})$
\begin{equation}\label{equ:14}
\begin{split}
D=\sum^{M}_{m=1}c_{m}\prod^{N}_{n=1}\frac{\partial^{T_{n}}}{\partial{x^{p_{mn}}_{n}}\partial{y^{q_{mn}}_{n}}}
\end{split}
\end{equation}
where $M$ denotes the number of terms, all coefficients $c_{m}$ are non-zero integers, $p_{mn}$ and $q_{mn}$ are non-negative integers, and positive integers $T_{n}$ represent the number of times that $(x_{n},y_{n})$ appears in the differential operator $D$.

As an instance, when setting $N=2$, $s_{11}=s_{12}=s_{22}=0$ and $t_{12}=2$, we can obtain
\begin{equation}\label{equ:15}
\begin{split}
D&=\left(\psi_{12}\right)^{2}\\&
=\left(\frac{\partial^{2}}{\partial{x_{1}}\partial{y_{2}}}-\frac{\partial^{2}}{\partial{x_{2}}\partial{y_{1}}}\right)^{2}\\&
=\frac{\partial^{2}}{\partial{x^{2}_{1}}}\frac{\partial^{2}}{\partial{y^{2}_{2}}}
-2\frac{\partial^{2}}{\partial{x_{1}}\partial{y_{1}}}\frac{\partial^{2}}{\partial{x_{2}}\partial{y_{2}}}
+\frac{\partial^{2}}{\partial{y^{2}_{1}}}\frac{\partial^{2}}{\partial{x^{2}_{2}}}\\&
\end{split}
\end{equation}
Obviously, it contains three terms, namely $M=3$, and $c_{1}=1$, $c_{2}=-2$ and $c_{3}=1$. Since both $(x_{1},y_{1})$ and $(x_{2},y_{2})$ are used twice in $D$, we have
\begin{equation}\label{equ:16}
\begin{split}
&(p_{11}+q_{11})=(p_{21}+q_{21})=(p_{31}+q_{31})=2\\&
(p_{12}+q_{12})=(p_{22}+q_{22})=(p_{32}+q_{32})=2
\end{split}
\end{equation}
\\
\\
\noindent\textbf{Definition 3.~} Suppose that an image function $f(x,y):\Omega\subset\mathbb{R}^{2}\rightarrow\mathbb{R}$ has infinite-order partial derivatives, we can define the function $RI$ as
\begin{equation}\label{equ:17}
RI=(D)(F)
\end{equation}
where
\begin{equation}\label{equ:18}
\begin{split}
F=\prod^{N}_{n=1}f(x_{n},y_{n})
\end{split}
\end{equation}
Note that each of $f(x_{n},y_{n})$ can be regarded as a copy of $f(x,y)$, which is derived by replacing the variable name $(x,y)$ with $f(x_{n},y_{n})$. Since changing the variable name does not influence the function itself, it is the same as $f(x,y)$.

Depending on (\ref{equ:14}), we can express $RI$ as
\begin{equation}\label{equ:19}
\begin{split}
RI&=(D)(F)\\&
=\left(\sum^{M}_{m=1}c_{m}\prod^{N}_{n=1}\frac{\partial^{T_{n}}}{\partial{x^{p_{mn}}_{n}}\partial{y^{q_{mn}}_{n}}}\right)
\left(\prod^{N}_{n=1}f(x_{n},y_{n})\right)\\&
=\sum^{M}_{m=1}c_{m}\prod^{N}_{n=1}\frac{\partial^{T_{n}}f(x_{n},y_{n})}{\partial{x^{p_{mn}}_{n}}\partial{y^{q_{mn}}_{n}}}
\end{split}
\end{equation}
Because all of $f(x_{n},y_{n})$ are copies of $f(x,y)$, they have identical partial derivatives, namely
\begin{equation}\label{equ:20}
\frac{\partial^{T_{n}}f(x_{n},y_{n})}{\partial{x^{p_{mn}}_{n}}\partial{y^{q_{mn}}_{n}}}
=\frac{\partial^{T_{n}}f(x,y)}{\partial{x^{p_{mn}}}\partial{y^{q_{mn}}}}
\end{equation}
Thus, (\ref{equ:19}) can be further simplified as
\begin{equation}\label{equ:21}
\begin{split}
RI&=\sum^{M}_{m=1}c_{m}\prod^{N}_{n=1}\frac{\partial^{T_{n}}f(x_{n},y_{n})}{\partial{x^{p_{mn}}_{n}}\partial{y^{q_{mn}}_{n}}}\\&
=\sum^{M}_{m=1}c_{m}\prod^{N}_{n=1}\frac{\partial^{T_{n}}f(x,y)}{\partial{x^{p_{mn}}}\partial{y^{q_{mn}}}}\\&
=\sum^{M}_{m=1}c_{m}\prod^{N}_{n=1}f_{p_{mn}q_{mn}}
\end{split}
\end{equation}
which means that $RI$ takes the form of homogeneous polynomial of $N$ degree in terms of partial derivatives of $f(x,y)$, and does not rely on any copies $f(x_{n},y_{n})$. For example, if we act the differential operator $D$ defined by (\ref{equ:15}) on $F=f(x_{1},y_{1})f(x_{2},y_{2})$, we can get
\begin{equation}\label{equ:22}
\begin{split}
RI&=\left(\psi_{12}\right)^{2}\left(f(x_{1},y_{1})f(x_{2},y_{2})\right)\\&
=2(f_{20}f_{02}-f^{2}_{11})
\end{split}
\end{equation}

The degree of $RI$ is equal to $N$, and the order of $RI$ is equal to $K=\max\limits_{m, n}\{p_{mn}+q_{mn}\}=\max\limits_{n}\{T_{n}\}$, namely the maximum number of times that $(x_{n},y_{n})$ used in the differential operator $D$. If two $RI$ have the same degree $N$, the same order $K$ and the same $T_{n}$ for all $n$, we say they have the same structure.
\\
\\
\noindent\textbf{Theorem 2.~} Suppose an image function $f(x,y):\Omega \subset\mathbb{R}^{2}\rightarrow \mathbb{R}$ has infinite-order partial derivatives. Let $f(x,y)$ be transformed using (\ref{equ:1}) into $h(u,v): \Omega ^{'}\subset\mathbb{R}^{2}\rightarrow \mathbb{R}$. We can define the function $RI^{'}$ as
\begin{equation}\label{equ:23}
\begin{split}
RI^{'}=(D^{'})(G)
\end{split}
\end{equation}
where
\begin{equation}\label{equ:24}
\begin{split}
&D^{'}=\prod_{1\leq i< j\leq N}(\psi^{'}_{ij})^{t_{ij}}
\prod_{1\leq i\leq j\leq N}(\phi^{'}_{ij})^{s_{ij}}\\&
G=\prod^{N}_{n=1}g(u_{n},v_{n})
\end{split}
\end{equation}
Then, we have $RI^{'}=RI$, where $RI$ is defined by (\ref{equ:17}).
\\
\\
\noindent\textbf{Proof.~} As stated previously, each of $h(u_{n},v_{n})$ is a copy of $h(u,v)$. It can be regarded as the rotated version of $f(x_{n},y_{n})$. Thus, we have
\begin{equation}\label{equ:25}
G=\prod^{N}_{n=1}g(u_{n},v_{n})=\prod^{N}_{n=1}f(x_{n},y_{n})=F
\end{equation}
where
\begin{equation}\label{equ:26}
\left(
  \begin{array}{c}
    u_{n} \\
    v_{n} \\
  \end{array}
\right)
=R_{\theta}
\left(
  \begin{array}{c}
    x_{n} \\
    y_{n} \\
  \end{array}
\right)
=
\left(
  \begin{array}{cc}
    \cos\theta & -\sin\theta \\
    \sin\theta & \cos\theta \\
  \end{array}
\right)
\left(
  \begin{array}{c}
    x_{n} \\
    y_{n} \\
  \end{array}
\right)
\end{equation}
Meanwhile, according to Theorem 1, we also have
\begin{equation}\label{equ:27}
\begin{split}
D^{'}&=\prod_{1\leq i< j\leq N}(\psi^{'}_{ij})^{t_{ij}}\prod_{1\leq i\leq j\leq N}(\phi^{'}_{ij})^{s_{ij}}\\&
=\prod_{1\leq i< j\leq N}(\psi_{ij})^{t_{ij}}\prod_{1\leq i\leq j\leq N}(\phi_{ij})^{s_{ij}}\\&
=D
\end{split}
\end{equation}
Substituting (\ref{equ:25}) and (\ref{equ:27}) into (\ref{equ:23}), we get
\begin{equation}\label{equ:28}
RI^{'}=(D^{'})(G)=(D)(F)=RI
\end{equation}
The proof is completed. \qed

This theorem shows that arbitrary $RI$ generated by (\ref{equ:17}) are rotation differential invariants of images.

\label{sec:7}
\begin{table*}
\caption{\label{tab1e:1} The differential operators used to generate rotation differential invariants $RI_{1}\sim RI_{230}$.}
\centering
\begin{tabular}{lll}
  \toprule
  \textbf{Differential Operators} & \textbf{Differential Operators}& \textbf{Differential Operators}\\
  \midrule
  $D_{1}=\phi_{11}$&
  $D_{2}=\phi_{12}$&
  $D_{3}=(\psi_{12})^{2}$\\
  \midrule
  $D_{4}=(\phi_{11})^{2}$&
  $D_{5}=\phi_{11}\phi_{22}$&
  $D_{6}=\phi_{12}\phi_{13}$\\
  \midrule
  $D_{7}=\phi_{12}\phi_{34}$&
  $D_{8}=\psi_{12}\psi_{13}$&
  $D_{9}=\phi_{11}\phi_{12}$\\
  \midrule
  $D_{10}=\phi_{11}\psi_{12}$&
  $D_{11}=\phi_{12}\psi_{13}$&
  $D_{12}=\phi_{11}\phi_{22}\phi_{33}$\\
  \midrule
  $D_{13}=(\phi_{12})^{3}$&
  $D_{14}=\phi_{12}\phi_{13}\phi_{14}$&
  $D_{15}=\phi_{12}\phi_{13}\phi_{23}$\\
  \midrule
  $D_{16}=\phi_{12}(\psi_{12})^{2}$&
  $D_{17}=\phi_{12}\psi_{13}\psi_{34}$&
  $D_{18}=\psi_{12}\psi_{13}\psi_{14}$\\
  \midrule
  $D_{19}=\phi_{11}\phi_{22}\phi_{33}\phi_{44}$&
  $D_{20}=(\phi_{12})^{4}$&
  $D_{21}=\phi_{11}\phi_{23}\psi_{12}\psi_{13}(\psi_{23})^{2}$\\
  \midrule
  $D_{22}=(\psi_{12})^{2}\psi_{13}\psi_{24}(\psi_{34})^{2}$&
  $D_{23}=(\phi_{11})^{2}\phi_{22}$&
  $D_{24}(\phi_{11})^{2}\phi_{23}$\\
  \midrule
  $D_{25}=\phi_{11}(\phi_{12})^{2}$&
  $D_{26}=\phi_{11}\phi_{12}\phi_{13}$&
  $D_{27}=\phi_{11}\phi_{22}\phi_{34}$\\
  \midrule
  $D_{28}=\phi_{11}\phi_{23}\phi_{24}$&
  $D_{29}=(\phi_{12})^{2}\phi_{13}$&
  $D_{30}=\phi_{11}\phi_{12}\psi_{12}$\\
  \midrule
  $D_{31}=\phi_{11}\phi_{12}\psi_{13}$&
  $D_{32}=\phi_{11}\phi_{23}\psi_{24}$&
  $D_{33}=(\phi_{12})^{2}\psi_{13}$\\
  \midrule
  $D_{34}=\phi_{12}\phi_{13}\psi_{14}$&
  $D_{35}=\phi_{12}(\psi_{13})^{2}$&
  $D_{36}=\phi_{12}\psi_{13}\psi_{14}$\\
  \midrule
  $D_{37}=(\psi_{12})^{2}\psi_{13}$&
  $D_{38}=(\phi_{11})^{2}(\phi_{22})^{2}$&
  $D_{39}=(\phi_{12})^{2}(\phi_{13})^{2}$\\
  \midrule
  $D_{40}=(\phi_{12})^{2}(\phi_{34})^{2}$&
  $D_{41}=\phi_{12}\phi_{13}\phi_{24}\phi_{34}$&
  $D_{42}=\phi_{12}\phi_{34}(\psi_{13})^{3}\psi_{24}$\\
  \midrule
  $D_{43}=\phi_{12}\phi_{34}(\psi_{13})^{3}(\psi_{24})^{3}$&
  $D_{44}=\phi_{11}\phi_{22}(\psi_{12})^{2}$&
  $D_{45}=\phi_{11}\phi_{12}\phi_{23}$\\
  \midrule
  $D_{46}=\phi_{11}\phi_{12}\psi_{23}$&
  $D_{47}=\phi_{11}(\phi_{23})^{3}$&
  $D_{48}=\phi_{11}(\phi_{23})^{2}\psi_{12}$\\
  \midrule
  $D_{49}=\phi_{11}\phi_{12}\psi_{23}\psi_{34}$&
  $D_{50}=(\phi_{11})^{2}\phi_{22}\phi_{33}$&
  $D_{51}=(\phi_{11})^{2}(\phi_{23})^{2}$\\
  \midrule
  $D_{52}=(\phi_{11})^{2}\phi_{23}\phi_{24}$&
  $D_{53}=(\phi_{12})^{2}\phi_{13}\phi_{14}$&
  $D_{54}=(\phi_{12})^{2}\phi_{13}\phi_{23}$\\
  \midrule
  $D_{55}=(\phi_{11})^{2}\psi_{23}\psi_{24}$&
  $D_{56}=(\phi_{12})^{2}(\phi_{34})^{2}\psi_{13}\psi_{24}$&
  $D_{57}=(\phi_{11})^{2}(\psi_{23})^{4}$\\
  \midrule
  $D_{58}=\phi_{11}\phi_{22}(\psi_{34})^{4}$&
  $D_{59}\phi_{11}\phi_{12}\phi_{33}\phi_{34}$&
  $D_{60}=(\phi_{12})^{2}\phi_{13}\phi_{34}$\\
  \midrule
  $D_{61}=\phi_{11}\phi_{12}\phi_{23}\psi_{24}$&
  $D_{62}=\phi_{11}\phi_{23}\psi_{12}\psi_{23}$&
  $D_{63}=(\phi_{12})^{4}\phi_{34}$\\
  \midrule
  $D_{64}=(\phi_{11})^{2}(\phi_{22})^{2}(\phi_{33})^{2}$&
  $D_{65}=(\phi_{12})^{3}(\phi_{34})^{3}$&
  $D_{66}=(\phi_{12})^{2}(\phi_{13})^{2}(\phi_{23})^{2}$\\
  \midrule
  $D_{67}=(\phi_{12})^{4}(\psi_{34})^{2}$&
  $D_{68}=\phi_{11}\phi_{22}\phi_{34}\psi_{13}\psi_{14}(\psi_{34})^{2}$&
  $D_{69}=\phi_{11}(\phi_{12})^{2}\phi_{33}$\\
  \midrule
  $D_{70}=\phi_{11}\phi_{12}(\phi_{23})^{2}$&
  $D_{71}=\phi_{11}\phi_{12}\psi_{23}\psi_{24}$&
  $D_{72}=(\phi_{11})^{2}\phi_{22}\phi_{23}$\\
  \midrule
  $D_{73}=\phi_{11}\phi_{12}\phi_{13}\phi_{44}$&
  $D_{74}=\phi_{11}\phi_{12}\phi_{22}\phi_{34}$&
  $D_{75}=\phi_{11}\phi_{12}\phi_{33}\phi_{44}$\\
  \midrule
  $D_{76}=\phi_{11}(\phi_{23})^{2}\phi_{24}$&
  $D_{77}=(\phi_{11})^{2}\phi_{22}\psi_{23}$&
  $D_{78}=\phi_{11}(\phi_{12})^{2}\psi_{23}$\\
  \midrule
  $D_{79}=\phi_{11}\phi_{12}\phi_{23}\psi_{12}$&
  $D_{80}=\phi_{11}\phi_{12}\phi_{33}\psi_{14}$&
  $D_{81}=(\phi_{12})^{2}\phi_{13}\psi_{14}$\\
  \midrule
  $D_{82}=(\phi_{12})^{2}\phi_{34}\psi_{13}$&
  $D_{83}=\phi_{11}\phi_{23}(\psi_{12})^{2}$&
  $D_{84}=(\phi_{11})^{2}(\phi_{22})^{2}\phi_{33}$\\
 \midrule
  $D_{85}=(\phi_{11})^{2}(\phi_{22})^{2}\phi_{34}$&
  $D_{86}=(\phi_{11})^{2}\phi_{22}\phi_{33}\phi_{44}$&
  $D_{87}=(\phi_{11})^{2}(\phi_{23})^{3}$\\
  \midrule
  $D_{88}=(\phi_{11})^{2}\phi_{23}\phi_{24}\phi_{34}$&
  $D_{89}=(\phi_{12})^{2}(\phi_{13})^{2}\phi_{23}$&
  $D_{90}=(\phi_{12})^{2}\phi_{13}(\phi_{34})^{2}$\\
  \midrule
  $D_{91}=(\phi_{11})^{2}\phi_{23}(\psi_{23})^{2}$&
  $D_{92}=\phi_{11}\phi_{12}\phi_{23}\psi_{12}\psi_{23}$&
  $D_{93}=\phi_{11}\phi_{12}\phi_{23}\psi_{12}\psi_{24}$\\
  \midrule
  $D_{94}=\phi_{11}\phi_{12}\phi_{23}\psi_{23}\psi_{34}$&
  $D_{95}=\phi_{11}\phi_{12}\phi_{33}\psi_{12}\psi_{23}$&
  $D_{96}=\phi_{11}(\phi_{23})^{2}\psi_{12}\psi_{34}$\\
 \midrule
  $D_{97}=\phi_{11}(\phi_{12})^{2}\phi_{23}\phi_{33}\psi_{23}$&
  $D_{98}=\phi_{11}\phi_{12}(\phi_{23})^{2}\phi_{44}\psi_{34}$&
  $D_{99}=(\phi_{11})^{2}(\phi_{22})^{2}(\psi_{34})^{2}$\\
  \midrule
  $D_{100}=\phi_{12}\phi_{13}\phi_{24}\phi_{34}(\psi_{14})^{2}$&
  $D_{101}=\phi_{11}\phi_{12}\phi_{34}(\psi_{23})^{3}$&
  $D_{102}\phi_{11}\phi_{12}(\psi_{23})^{3}\psi_{34}$\\
  \midrule
  $D_{103}=(\psi_{12})^{2}\psi_{13}\psi_{14}\psi_{23}\psi_{24}$&
  $D_{104}=(\phi_{12})^{3}(\psi_{34})^{4}$&
  $D_{105}=\phi_{12}\phi_{13}\phi_{24}\psi_{12}(\psi_{34})^{3}$\\
  \midrule
  $D_{106}=\phi_{11}\phi_{12}\phi_{33}\psi_{23}$&
  $D_{107}=(\phi_{12})^{2}\phi_{13}\phi_{23}\phi_{34}$&
  $D_{108}=\phi_{12}\phi_{13}\phi_{14}\phi_{23}\phi_{24}$\\
 \midrule
  $D_{109}=\phi_{11}\phi_{12}\phi_{23}\phi_{34}\psi_{24}$&
  $D_{110}=\phi_{12}\phi_{13}\phi_{14}\phi_{23}\psi_{14}$&
  $D_{111}=\phi_{11}\phi_{12}\phi_{33}\phi_{34}(\psi_{24})^{2}$\\
  \midrule
  $D_{112}=\phi_{11}\phi_{12}\phi_{13}\phi_{24}$&
  $D_{113}=\phi_{11}\phi_{22}(\phi_{34})^{4}$&
  $D_{114}=(\phi_{11})^{2}(\phi_{22})^{2}(\phi_{33})^{2}(\phi_{44})^{2}$\\
  \midrule
  $D_{115}=(\phi_{12})^{4}(\phi_{34})^{4}$&
  $D_{116}=\phi_{11}\phi_{12}(\phi_{23})^{2}\phi_{34}\phi_{44}\psi_{13}\psi_{24}$&
  $D_{117}=(\phi_{12})^{4}(\psi_{34})^{4}$\\
\bottomrule
\end{tabular}
\label{table:1}
\end{table*}

\begin{table*}
\caption{\label{tab1e:2} The differential operators used to generate rotation differential invariants $RI_{1}\sim RI_{230}$. (Continued)}
\centering
\begin{tabular}{lll}
  \toprule
  \textbf{Differential Operators} & \textbf{Differential Operators} & \textbf{Differential Operators} \\
  \midrule
  $D_{118}=\phi_{11}\phi_{12}\phi_{23}\psi_{34}$&
  $D_{119}=\phi_{11}\phi_{22}\phi_{34}\psi_{13}$&
  $D_{120}=\phi_{11}\phi_{22}\psi_{13}\psi_{34}$\\
  \midrule
  $D_{121}=(\phi_{11})^{2}\phi_{22}(\phi_{23})^{2}$&
  $D_{122}=(\phi_{11})^{2}\phi_{22}\phi_{23}\phi_{24}$&
  $D_{123}=\phi_{11}(\phi_{12})^{2}\phi_{22}\phi_{33}$\\
  \midrule
  $D_{124}=\phi_{11}(\phi_{12})^{2}\phi_{22}\phi_{34}$&
  $D_{125}=\phi_{11}(\phi_{12})^{2}\phi_{33}\phi_{44}$&
  $D_{126}=(\phi_{11})^{2}\phi_{22}\phi_{23}\psi_{23}$\\
  \midrule
  $D_{127}=\phi_{11}\phi_{12}\phi_{23}\phi_{34}\psi_{14}$&
  $D_{128}=\phi_{11}(\phi_{12})^{2}\phi_{23}\phi_{34}\psi_{24}$&
  $D_{129}=\phi_{11}\phi_{12}(\phi_{23})^{2}\phi_{44}\psi_{12}$\\
  \midrule
  $D_{130}=\phi_{11}\phi_{12}(\phi_{23})^{2}\phi_{34}\phi_{44}\psi_{12}\psi_{34}$&
  $D_{131}=(\phi_{11})^{2}(\phi_{22})^{2}(\psi_{34})^{4}$&
  $D_{132}=(\phi_{11})^{2}(\phi_{23})^{2}\phi_{24}$\\
  \midrule
  $D_{133}=\phi_{11}(\phi_{12})^{2}(\phi_{34})^{2}$&
  $D_{134}=\phi_{11}\phi_{12}\phi_{22}\phi_{33}\phi_{44}$&
  $D_{135}=\phi_{11}\phi_{12}(\phi_{34})^{2}\psi_{12}$\\
  \midrule
  $D_{136}=(\phi_{11})^{2}(\phi_{22})^{2}\phi_{33}\phi_{44}$&
  $D_{137}=(\phi_{11})^{2}(\phi_{23})^{2}(\phi_{24})^{2}$&
  $D_{138}=(\phi_{11})^{2}\phi_{23}\phi_{24}(\psi_{34})^{2}$\\
  \midrule
  $D_{139}=(\phi_{12})^{2}\phi_{13}\phi_{14}\psi_{23}\psi_{34}$&
  $D_{140}=(\phi_{12})^{2}\phi_{13}\phi_{34}\psi_{14}\psi_{23}$&
  $D_{141}=(\phi_{12})^{2}\phi_{13}\phi_{34}(\psi_{23})^{2}$\\
  \midrule
  $D_{142}=(\phi_{12})^{2}\phi_{13}(\psi_{23})^{2}\psi_{34}$&
  $D_{143}=(\phi_{11})^{2}(\psi_{23})^{2}(\psi_{24})^{2}$&
  $D_{144}=(\phi_{11})^{2}(\psi_{23})^{2}\psi_{24}\psi_{34}$\\
  \midrule
  $D_{145}=(\phi_{12})^{2}\psi_{13}\psi_{14}(\psi_{34})^{2}$&
  $D_{146}=\phi_{12}(\psi_{13})^{2}\psi_{14}(\psi_{34})^{2}$&
  $D_{147}=(\psi_{12})^{2}(\psi_{13})^{2}\psi_{23}\psi_{24}$\\
  \midrule
  $D_{148}=\phi_{11}\phi_{12}\phi_{13}\phi_{24}(\psi_{34})^{3}$&
  $D_{149}=(\phi_{11})^{2}\phi_{22}(\psi_{34})^{4}$&
  $D_{150}=\phi_{11}(\psi_{12})^{2}(\psi_{34})^{4}$\\
  \midrule
  $D_{151}=\phi_{11}(\phi_{12})^{2}\phi_{23}\phi_{33}$&
  $D_{152}=\phi_{11}\phi_{12}(\phi_{23})^{2}\phi_{44}$&
  $D_{153}=(\phi_{12})^{2}\phi_{13}\phi_{24}\psi_{14}$\\
  \midrule
  $D_{154}=\phi_{11}(\phi_{23})^{2}\psi_{14}\psi_{24}$&
  $D_{155}=\phi_{11}\phi_{23}\phi_{24}\psi_{13}\psi_{24}$&
  $D_{156}=\phi_{11}\phi_{12}(\phi_{34})^{4}$\\
  \midrule
  $D_{157}=(\phi_{12})^{2}(\phi_{13})^{2}\phi_{24}\phi_{34}$&
  $D_{158}=\phi_{11}(\phi_{23})^{4}\psi_{14}$&
  $D_{159}=(\phi_{12})^{2}(\phi_{13})^{2}\psi_{24}\psi_{34}$\\
  \midrule
  $D_{160}=\phi_{12}\phi_{13}(\psi_{24})^{2}(\psi_{34})^{2}$&
  $D_{161}=(\psi_{12})^{2}(\psi_{13})^{2}\psi_{24}\psi_{34}$&
  $D_{162}=(\phi_{11})^{2}(\phi_{22})^{2}(\phi_{33})^{2}\phi_{44}$\\
  \midrule
  $D_{163}=\phi_{11}(\phi_{23})^{2}(\phi_{24})^{2}(\phi_{34})^{2}$&
  $D_{164}=(\phi_{12})^{4}(\phi_{34})^{3}$&
  $D_{165}=(\phi_{12})^{3}\phi_{13}(\phi_{34})^{3}$\\
  \midrule
  $D_{166}=(\phi_{12})^{4}\phi_{34}(\psi_{34})^{2}$&
  $D_{167}=\phi_{11}\phi_{12}\phi_{23}(\phi_{34})^{2}\psi_{13}$&
  $D_{168}=(\phi_{11})^{2}\phi_{22}\phi_{23}\phi_{34}$\\
  \midrule
  $D_{169}=(\phi_{11})^{2}\phi_{22}\phi_{23}\phi_{44}$&
  $D_{170}=\phi_{11}(\phi_{12})^{2}\phi_{23}\phi_{34}$&
  $D_{171}=\phi_{11}(\phi_{12})^{2}\phi_{23}\phi_{44}$\\
  \midrule
  $D_{172}=\phi_{11}(\phi_{12})^{2}\phi_{33}\phi_{34}$&
  $D_{173}=\phi_{11}\phi_{12}\phi_{13}(\phi_{24})^{2}$&
  $D_{174}=\phi_{11}\phi_{12}(\phi_{23})^{2}\phi_{24}$\\
  \midrule
  $D_{175}=(\phi_{11})^{2}\phi_{22}\phi_{23}\psi_{34}$&
  $D_{176}=(\phi_{11})^{2}\phi_{22}\phi_{33}\psi_{24}$&
  $D_{177}=(\phi_{12})^{2}\phi_{13}\phi_{34}\psi_{12}$\\
  \midrule
  $D_{178}=(\phi_{12})^{2}\phi_{13}\phi_{34}\psi_{13}$&
  $D_{179}=\phi_{11}\phi_{12}\phi_{13}(\psi_{24})^{2}$&
  $D_{180}=(\phi_{11})^{2}(\phi_{22})^{2}\phi_{33}\phi_{34}$\\
  \midrule
  $D_{181}=(\phi_{11})^{2}\phi_{22}(\phi_{34})^{3}$&
  $D_{182}=(\phi_{11})^{2}(\phi_{23})^{3}\phi_{24}$&
  $D_{183}=\phi_{11}(\phi_{12})^{2}(\phi_{34})^{3}$\\
  \midrule
  $D_{184}=(\phi_{12})^{3}\phi_{13}\phi_{23}\phi_{34}$&
  $D_{185}=(\phi_{11})^{2}(\phi_{22})^{2}\phi_{33}\psi_{34}$&
  $D_{186}=(\phi_{11})^{2}\phi_{22}(\phi_{34})^{2}\psi_{23}$\\
  \midrule
  $D_{187}=(\phi_{11})^{2}(\phi_{23})^{3}\psi_{24}$&
  $D_{188}=\phi_{11}(\phi_{12})^{2}\phi_{23}\phi_{34}\psi_{23}$&
  $D_{189}=\phi_{11}(\phi_{12})^{2}\phi_{33}\phi_{44}\psi_{23}$\\
  \midrule
  $D_{190}=\phi_{11}(\phi_{12})^{2}(\phi_{34})^{2}\psi_{23}$&
  $D_{191}=\phi_{11}\phi_{12}(\phi_{23})^{2}\phi_{34}\psi_{12}$&
  $D_{192}=\phi_{11}\phi_{12}(\phi_{34})^{3}\psi_{12}$\\
  \midrule
  $D_{193}=\phi_{11}(\phi_{23})^{2}(\phi_{24})^{2}\psi_{13}$&
  $D_{194}=(\phi_{12})^{3}\phi_{13}\phi_{23}\psi_{34}$&
  $D_{195}=(\phi_{12})^{3}(\phi_{34})^{2}\psi_{13}$\\
  \midrule
  $D_{196}=\phi_{11}(\phi_{12})^{2}\phi_{23}\psi_{23}\psi_{34}$&
  $D_{197}=\phi_{11}\phi_{12}(\phi_{23})^{2}\psi_{12}\psi_{34}$&
  $D_{198}=(\phi_{12})^{3}\psi_{13}(\psi_{34})^{2}$\\
  \midrule
  $D_{199}=(\phi_{12})^{2}\phi_{13}(\psi_{34})^{3}$&
  $D_{200}=\phi_{12}(\psi_{13})^{3}(\psi_{24})^{2}$&
  $D_{201}=(\phi_{11})^{2}(\phi_{22})^{2}(\phi_{34})^{3}$\\
  \midrule
  $D_{202}=\phi_{11}(\phi_{12})^{2}(\phi_{23})^{2}\phi_{44}\psi_{34}$&
  $D_{203}=\phi_{11}(\phi_{12})^{2}\phi_{23}\phi_{33}\phi_{44}\psi_{23}$&
  $D_{204}=\phi_{11}\phi_{12}\phi_{13}\phi_{23}(\phi_{24})^{2}\psi_{34}$\\
  \midrule
  $D_{205}=(\phi_{11})^{2}(\phi_{22})^{2}\phi_{34}(\psi_{34})^{2}$&
  $D_{206}=\phi_{11}(\phi_{12})^{2}\phi_{23}\phi_{44}\psi_{23}\psi_{34}$&
  $D_{207}=\phi_{11}\phi_{12}(\phi_{23})^{2}\phi_{44}\psi_{12}\psi_{34}$\\
  \midrule
  $D_{208}=(\phi_{12})^{2}\phi_{13}\phi_{24}\phi_{34}\psi_{14}\psi_{34}$&
  $D_{209}=(\phi_{12})^{3}\psi_{13}(\psi_{34})^{3}$&
  $D_{210}=\phi_{11}\phi_{12}\phi_{13}\phi_{23}\phi_{24}\psi_{24}$\\
  \midrule
  $D_{211}=\phi_{11}(\phi_{12})^{2}\phi_{23}(\phi_{34})^{2}\psi_{23}$&
  $D_{212}=\phi_{11}\phi_{12}(\phi_{23})^{2}(\phi_{34})^{2}\psi_{12}$&
  $D_{213}=(\phi_{12})^{3}\phi_{13}(\psi_{34})^{3}$\\
  \midrule
  $D_{214}=(\phi_{11})^{2}\phi_{22}(\phi_{23})^{2}\phi_{34}\phi_{44}\psi_{34}$&
  $D_{215}=(\phi_{11})^{2}(\phi_{23})^{2}\phi_{24}\phi_{34}$&
  $D_{216}=(\phi_{12})^{2}\phi_{13}\phi_{14}(\phi_{34})^{2}$\\
  \midrule
  $D_{217}=(\phi_{12})^{3}\phi_{13}\phi_{34}\psi_{34}$&
  $D_{218}=\phi_{11}(\phi_{12})^{2}\phi_{23}(\phi_{34})^{2}\psi_{24}$&
  $D_{219}=\phi_{11}\phi_{12}(\phi_{23})^{2}\phi_{34}\phi_{44}\psi_{12}$\\
  \midrule
  $D_{220}=\phi_{11}\phi_{12}\phi_{13}\phi_{24}\phi_{34}(\psi_{23})^{2}$&
  $D_{221}=(\phi_{12})^{2}\phi_{13}\phi_{14}\phi_{23}\psi_{23}\psi_{34}$&
  $D_{222}=(\phi_{12})^{2}\phi_{13}\phi_{24}\phi_{34}\psi_{14}\psi_{23}$\\
  \midrule
  $D_{223}=(\phi_{12})^{2}\phi_{13}(\phi_{34})^{2}\psi_{14}\psi_{23}$&
  $D_{224}=(\phi_{11})^{2}\phi_{22}\phi_{23}\phi_{34}\psi_{34}$&
  $D_{225}=(\phi_{11})^{2}(\phi_{22})^{2}\phi_{33}(\phi_{34})^{2}$\\
  \midrule
  $D_{226}=\phi_{11}\phi_{12}\phi_{22}(\phi_{34})^{3}\psi_{13}$&
  $D_{227}=\phi_{11}\phi_{12}(\phi_{34})^{4}\psi_{12}$&
  $D_{228}=\phi_{11}\phi_{12}\phi_{33}\phi_{34}\psi_{14}$\\
  \midrule
  $D_{229}=\phi_{11}\phi_{12}\phi_{33}\phi_{44}\psi_{13}$&
  $D_{230}=\phi_{11}(\phi_{12})^{2}\phi_{23}(\phi_{34})^{2}\psi_{34}$&\\
  \bottomrule
\end{tabular}
\label{table:2}
\end{table*}

\section{The Dependencies among Rotation Differential Invariants}
\label{sec:5}
When setting the degree $N\leq 4$ and the order $K\leq 4$, we generate all possible $RI$ using (\ref{equ:17}). Our method does not guarantee that there are no dependent invariants in this generated set, which means some of $RI$ is the function of the others. In many practice applications, a single real-valued $RI$ does not provide enough discrimination power, and all of $RI$ in the set must be used simultaneously as a invariant vector. Previous researches have found that dependent differential invariants do not contribute to the discrimination power of the vector, they only increase the dimensionality of feature space. This leads not only to the growth of the time complexity, but also to a drop of performance. Thus, identifying and discarding dependent $RI$ in the derived set is highly desirable.

Let us look at the possible dependencies among $RI$ in detail. We categorize them into three groups and explain how they can be eliminated.

\subsection{Linear Dependencies}
\label{sec:5.1}
Some $RI$ in the generated set may be a linear combination of the others. It is obvious that only $RI$ of the same structure can be linearly dependent. Thus, we first group all $RI$ according to their structure. Then, for each group, a matrix of coefficients of all $RI$ can be derived. The "coefficient" here means the multiplicative constant of each term, namely $c_{m}$ in (\ref{equ:21}). If a $RI$ does not contain all possible terms, the corresponding coefficients are zero. This ensures that all coefficient vectors are of the same length, and we can arrange them into a matrix. For example, when setting $N=2$, $K=2$ and $T_{1}=T_{2}=2$, the invariants with the same structure may contain six terms $\{f^{2}_{20}, f_{20}f_{11},f_{20}f_{02},f^{2}_{11},f_{11}f_{02},f^{2}_{02}\}$. Hence, the coefficient vector of $RI=2(f_{20}f_{02}-f^{2}_{11})$ is $(0,0,2,2,0,0)$.

If the derived matrix has a full rank, all $RI$ are linearly independent. Some symbolic computing softwares are used to derive the coefficient matrix of each group and to calculate its rank. We finally derive two hundred and thirty invariants from the generated set, which are linearly independent. In the following, they are denoted as $RI_{1}\sim RI_{230}$, and the set of them is denoted as $S_{L}(4, 4)$.

The differential operators $D_{i}$ listed in Table.~\ref{table:1} and Table.~\ref{table:2} can be used to generate the corresponding $RI_{i}$, where $i=1,2,...,230$. We can find that only $(x_{1},y_{1})$, $(x_{2},y_{2})$, $(x_{3},y_{3})$ and $(x_{4},y_{4})$ are involved in each $D_{i}$, and each of them appears at most four times in $D_{i}$. This guarantees the degree $N$ and the order $K$ of $RI_{i}$ are less than or equal to $4$. For example, we have
\begin{equation}\label{equ:29}
\begin{split}
RI_{2}&=(D_{2})(F)\\&
=(\phi_{12})\left(f(x_{1},y_{1})f(x_{2},y_{2})\right)\\&
=\left(\frac{\partial^{2}}{\partial{x_{1}}\partial{x_{2}}}+\frac{\partial^{2}}{\partial{y_{1}}\partial{y_{2}}}\right)\left(f(x_{1},y_{1})f(x_{2},y_{2})\right)\\&
=f^{2}_{10}+f^{2}_{01}\\
\end{split}
\end{equation}

\subsection{Polynomial Dependencies}
\label{sec:5.2}
After linear dependent invariants have been eliminated, there may still polynomial dependencies among $RI_{1}\sim RI_{230}$, which means some of them can be expressed as a linear combination of the other invariants and their products.

All possible products of invariants can be generated by $P=\prod^{230}_{i=1}(RI_{i})^{n_{i}}$, where $n_{i}$ are non-negative integers. They are also homogeneous polynomials in partial derivatives $f_{pq}$. If the degree of $P$ is greater than $4$, $P$ and all $RI_{i}$ must be linearly independent, because they have different structures. Thus, we only generate $P$ whose degree is less than $4$, such as $(RI_{1})^{2}=f^{4}_{10}+f^{4}_{01}$, and add them to the set $S_{L}(4, 4)$. Then, linear dependencies among all invariants in this larger set are derived.

We discover one hundred and thirty-four polynomially dependent relations among $RI_{1}\sim RI_{230}$. They are listed in Table.~\ref{table:3} and Table.~\ref{table:4}. Thus, only ninety-six invariants in the set $S_{L}(4, 4)$ are polynomially independent. The set of them is denoted as $S_{P}(4, 4)$.

\begin{table*}
\newcommand{\tabincell}[2]{\begin{tabular}{@{}#1@{}}#2\end{tabular}}
\renewcommand
\arraystretch{1.12}
\caption{The polynomial dependencies among rotation differential invariants $RI_{1}\sim RI_{230}$.}
\centering
\begin{tabular}{ll}
  \toprule
  \textbf{Polynomial Relation} &\textbf{Polynomial Relation}\\
  \midrule
  $RI_{5}=RI^{2}_{1}$ & $RI_{7}=RI^{2}_{2}$\\
  \midrule
  $RI_{8}=RI_{1}RI_{2}-RI_{6}$ & $RI_{12}=RI^{3}_{1}$\\
  \midrule
  $RI_{15}=RI^{3}_{1}-\frac{3}{2}RI_{1}RI_{3}$ & $RI_{17}=-\frac{1}{2}RI_{2}RI_{3}$\\
  \midrule
  $RI_{19}=RI^{4}_{1}$ & $RI_{23}=RI_{1}RI_{4}$\\
  \midrule
  $RI_{24}=RI_{2}RI_{4}$ & $RI_{27}=RI^{2}_{1}RI_{2}$\\
  \midrule
  $RI_{28}=RI_{1}RI_{6}$ & $RI_{32}=RI_{1}RI_{11}$\\
  \midrule
  $RI_{34}=RI_{2}RI_{10}-RI_{18}$ & $RI_{35}=RI_{1}RI_{9}-RI_{29}$\\
  \midrule
  $RI_{36}=RI_{2}RI_{9}-RI_{14}$ & $RI_{37}=RI_{1}RI_{10}-RI_{33}$\\
  \midrule
  $RI_{38}=RI^{2}_{4}$ & $RI_{40}=RI^{2}_{3}+RI^{2}_{1}(RI^{2}_{1}-2RI_{3})$\\
  \midrule
  $RI_{41}=RI^{2}_{1}(RI^{2}_{1}-2RI_{3})+\frac{1}{2}RI^{2}_{3}$ &$RI_{42}=\frac{1}{2}RI_{3}(2RI_{44}+RI_{20}-RI^{2}_{4})$\\
  \midrule
  \tabincell{l}{$RI_{43}=2RI_{44}(RI_{44}-RI^{2}_{4})+RI_{20}(2RI_{44}+\frac{1}{2}RI_{20}$\\$~~~~~~~~~-RI^{2}_{4})+\frac{1}{2}RI^{4}_{4}$} & $RI_{47}=RI_{1}RI_{13}$\\
  \midrule
  $RI_{49}=-\frac{1}{2}RI_{3}RI_{9}$ & $RI_{50}=RI^{2}_{1}RI_{4}$\\
  \midrule
  $RI_{51}=RI_{4}RI_{5}-RI_{3}RI_{4}$ & $RI_{52}=RI_{4}RI_{6}$\\
  \midrule
  $RI_{55}=RI_{4}(RI_{1}RI_{2}-RI_{6})$ & $RI_{56}=RI^{2}_{16}-RI_{22}$\\
  \midrule
  $RI_{57}=RI_{4}(RI_{20}+2RI_{44}-RI^{2}_{4})$ & $RI_{58}=RI^{2}_{1}(2RI_{44}+RI_{20}-RI^{2}_{4})$\\
  \midrule
  $RI_{59}=RI^{2}_{9}$ & $RI_{63}=RI_{2}RI_{20}$\\
  \midrule
  $RI_{64}=RI^{3}_{4}$ & $RI_{65}=RI^{2}_{13}$\\
  \midrule
  $RI_{66}=\frac{1}{2}RI_{4}(3RI_{20}-RI^{2}_{4})+\frac{3}{2}RI_{21}$ & $RI_{67}=RI_{3}RI_{20}$\\
  \midrule
  $RI_{68}=RI_{1}RI_{21}$ & $RI_{69}=RI_{1}RI_{25}$\\
  \midrule
  $RI_{70}=\frac{1}{2}RI_{1}RI_{16}+RI_{54}$ & $RI_{72}=RI_{4}RI_{9}$\\
  \midrule
  $RI_{73}=RI_{1}RI_{26}$ & $RI_{74}=RI^{2}_{9}+RI^{2}_{10}$\\
  \midrule
  $RI_{75}=RI^{2}_{1}RI_{9}$ & $RI_{76}=RI_{1}RI_{29}$\\
  \midrule
  $RI_{77}=RI_{4}RI_{10}$ & $RI_{84}=RI^{2}_{4}RI_{1}$\\
  \midrule
  $RI_{85}=RI^{2}_{4}RI_{2}$ & $RI_{86}=RI^{3}_{1}RI_{4}$\\
  \midrule
  $RI_{87}=RI_{4}RI_{13}$ & $RI_{88}=RI_{1}RI_{4}(RI^{2}_{1}-\frac{3}{2}RI_{3})$\\
  \midrule
  $RI_{91}=RI_{4}RI_{16}$ & $RI_{94}=-\frac{1}{2}RI_{9}RI_{16}$\\
  \midrule
  $RI_{99}=RI^{2}_{4}RI_{3}$ & \tabincell{l}{$RI_{100}=\frac{1}{4}RI^{2}_{1}(RI^{2}_{4}-RI_{44}-RI_{20})+\frac{1}{4}RI_{4}(2RI_{1}RI_{25}$\\~~~~~~~~~~~~$-RI_{4}RI_{3}-2RI_{39})-RI_{1}RI_{92}-\frac{1}{2}RI^{2}_{30}$}\\
  \midrule
  $RI_{101}=\frac{1}{2}RI_{10}(2RI_{44}+RI_{20}-RI^{2}_{4})$ & $RI_{102}=\frac{1}{2}RI_{9}(RI^{2}_{4}-RI_{20}-2RI_{44})$\\
  \midrule
  \tabincell{l}{$RI_{103}=\frac{1}{4}RI^{2}_{1}(RI^{2}_{4}-RI_{44}-RI_{20})+\frac{1}{4}RI_{4}(2RI_{1}RI_{25}$\\~~~~~~~~~~~~$-RI_{4}RI_{3}-2RI_{39})+RI_{1}RI_{92}-\frac{1}{2}RI^{2}_{30}$}
  & $RI_{104}=RI_{13}(2RI_{44}+RI_{20}-RI^{2}_{4})$\\
  \midrule
  $RI_{105}=\frac{1}{2}RI_{16}(2RI_{44}-RI^{2}_{4}+RI_{20})$ & $RI_{107}=RI_{96}-\frac{1}{2}RI_{9}(RI_{16}-2RI_{13})$\\
  \midrule
  \tabincell{l}{$RI_{108}=RI_{90}-\frac{1}{2}RI_{3}RI_{16}$} & $RI_{109}=-RI_{1}(RI_{48}+RI_{106})$\\
  \midrule
  $RI_{111}=\frac{1}{2}RI_{16}(2RI_{13}+3RI_{16})-RI_{22}$ & $RI_{112}=\frac{1}{2}RI_{2}RI_{25}+\frac{1}{2}RI_{1}RI_{26}-\frac{1}{2}RI_{4}(RI_{1}RI_{2}-RI_{6})$\\
  \midrule
  $RI_{113}=RI^{2}_{1}RI_{20}$ & $RI_{114}=RI^{4}_{4}$\\
  \midrule
  $RI_{115}=RI^{2}_{20}$ & $RI_{116}=\frac{1}{2}RI_{4}RI_{21}$\\
  \midrule
  $RI_{117}=RI_{20}(2RI_{44}+RI_{20}-RI^{2}_{4})$ & $RI_{120}=RI_{1}(RI_{45}-RI_{1}RI_{9})$\\
  \midrule
  $RI_{121}=RI_{4}RI_{25}$ & $RI_{122}=RI_{4}RI_{26}$\\
  \midrule
  $RI_{123}=RI_{1}(RI^{2}_{4}-RI_{44})$ & $RI_{124}=RI_{2}(RI^{2}_{4}-RI_{44})$\\
  \bottomrule
\end{tabular}
\label{table:3}
\end{table*}

\begin{table*}
\newcommand{\tabincell}[2]{\begin{tabular}{@{}#1@{}}#2\end{tabular}}
\renewcommand
\arraystretch{1.2}
\caption{The polynomial dependencies among rotation differential invariants $RI_{1}\sim RI_{230}$.(Continued)}
\centering
\begin{tabular}{ll}
  \toprule
  \textbf{Polynomial Dependencies} &\textbf{Polynomial Dependencies}\\
  \midrule
  $RI_{125}=RI^{2}_{1}RI_{25}$ & $RI_{126}=RI_{4}RI_{30}$\\
  \midrule
  $RI_{127}=RI_{30}(RI^{2}_{1}-\frac{1}{2}RI_{3})$ & $RI_{129}=RI_{1}RI_{4}RI_{30}-RI_{128}$\\
  \midrule
  $RI_{130}=\frac{1}{2}RI_{44}(RI_{44}-RI^{2}_{4}+RI_{20})+\frac{1}{2}RI_{4}RI_{21}$ & $RI_{131}=RI^{2}_{4}(2RI_{44}+RI_{20}-RI^{2}_{4})$\\
  \midrule
  $RI_{132}=RI_{4}RI_{29}$ & $RI_{133}=RI_{25}(RI^{2}_{1}-RI_{3})$\\
  \midrule
  $RI_{134}=RI^{2}_{1}(RI_{13}+RI_{16})$ & $RI_{135}=RI_{30}(RI^{2}_{1}-RI_{3})$\\
  \midrule
  $RI_{136}=RI^{2}_{1}RI^{2}_{4}$ & $RI_{137}=RI_{4}RI_{39}$\\
  \midrule
  $RI_{138}=RI_{4}(\frac{1}{2}RI_{1}RI_{16}-RI_{62})$ & $RI_{140}=RI_{139}-\frac{1}{2}RI_{1}(RI_{95}-\frac{1}{2}RI_{4}RI_{16})$\\
  \midrule
  \tabincell{l}{$RI_{143}=RI^{2}_{1}(RI^{2}_{4}-RI_{44})+RI_{3}(2RI_{44}-RI^{2}_{4})$\\$~~~~~~~~~~~~+RI_{4}RI_{39}-2RI^{2}_{25}-2RI^{2}_{30}$} & $RI_{144}=RI_{4}(RI_{62}+\frac{1}{2}RI_{1}RI_{16})$\\
  \midrule
  $RI_{145}=RI_{16}RI_{25}+2RI_{139}$ & $RI_{149}=RI_{1}RI_{4}(2RI_{44}+RI_{20}-RI^{2}_{4})$\\
  \midrule
  $RI_{150}=RI_{25}(2RI_{44}+RI_{20}-RI^{2}_{4})$ & $RI_{152}=RI_{1}(RI_{54}+\frac{1}{2}RI_{1}RI_{16})$\\
  \midrule
  \tabincell{l}{$RI_{155}=RI_{154}-\frac{1}{2}RI_{3}(RI_{13}+RI_{16})$} & $RI_{156}=RI_{9}RI_{20}$\\
  \midrule
  $RI_{158}=RI_{10}RI_{20}$ & $RI_{159}=RI_{1}RI_{89}-RI_{157}$\\
  \midrule
  $RI_{161}=-RI_{160}+RI_{1}(RI_{4}RI_{16}+RI_{89}+RI_{4}RI_{13}-2RI_{151})$ & $RI_{162}=RI_{1}RI^{3}_{4}$\\
  \midrule
  $RI_{163}=\frac{1}{2}RI_{1}(3RI_{4}RI_{20}+3RI_{21}-RI^{3}_{4})$ & $RI_{164}=RI_{13}RI_{20}$\\
  \midrule
  $RI_{166}=RI_{16}RI_{20}$ & $RI_{168}=RI_{4}RI_{45}$\\
  \midrule
  $RI_{169}=RI_{1}RI_{4}RI_{9}$ & $RI_{171}=RI_{1}(RI_{4}RI_{9}-RI_{83})$\\
  \midrule
  $RI_{172}=RI_{9}RI_{25}$ & $RI_{175}=RI_{4}RI_{46}$\\
  \midrule
  $RI_{176}=RI_{1}RI_{4}RI_{10}$ & \tabincell{l}{$RI_{179}=RI_{10}RI_{30}+RI_{4}(RI_{45}-RI_{29})$\\$~~~~~~~~~~~~+RI_{1}(RI_{4}RI_{9}-RI_{83})-2RI_{170}$\\$~~~~~~~~~~~~+RI_{173}$}\\
  \midrule
  $RI_{180}=RI^{2}_{4}RI_{9}$ & $RI_{181}=RI_{1}RI_{4}RI_{13}$\\
  \midrule
  $RI_{183}=RI_{13}RI_{25}$ & $RI_{185}=RI^{2}_{4}RI_{10}$\\
  \midrule
  $RI_{186}=RI_{4}RI_{48}$ & $RI_{190}=\frac{1}{2}(RI_{16}RI_{30}+RI_{4}RI_{48}+RI_{189})$\\
  \midrule
  $RI_{191}=RI_{4}RI_{79}-RI_{188}$ & $RI_{192}=RI_{13}RI_{30}$\\
  \midrule
  \tabincell{l}{$RI_{194}=\frac{1}{2}[-RI_{142}-RI_{146}-RI_{188}+RI_{4}(2RI_{78}$\\$~~~~~~~~~~~~+RI_{79})+RI_{10}(RI_{20}-RI^{2}_{4})]$} & \tabincell{l}{$RI_{197}=-RI_{141}+RI_{147}-2RI_{184}+RI_{9}RI_{20}$\\$~~~~~~~~~~~~+RI_{4}(RI_{4}RI_{9}-2RI_{83})$} \\
  \midrule
  $RI_{201}=RI^{2}_{4}RI_{13}$ & $RI_{203}=RI_{1}RI_{97}$\\
  \midrule
  $RI_{205}=RI^{2}_{4}RI_{16}$ & $RI_{207}=RI_{4}RI_{95}-RI_{206}$\\
  \midrule
  $RI_{210}=RI_{25}RI_{30}-RI_{167}$ & $RI_{212}=2RI_{148}+RI_{211}-RI_{1}RI_{97}-RI_{30}RI_{44}$ \\
  \midrule
  $RI_{214}=RI_{4}RI_{97}$ & $RI_{215}=RI_{4}RI_{54}$\\
  \midrule
  $RI_{221}=\frac{1}{4}RI_{1}RI_{21}+\frac{1}{2}RI_{25}(RI_{44}+RI_{20}-RI^{2}_{4})$ & \tabincell{l}{$RI_{222}=-\frac{2}{3}RI_{209}+RI_{4}(RI_{151}-RI_{89})$\\$~~~~~~~~~~~~-\frac{1}{6}RI_{13} (RI_{44}-RI_{20}+RI^{2}_{4})$\\$~~~~~~~~~~~~+\frac{1}{2}RI_{16}(RI_{44}-2RI^{2}_{4}+RI_{20})$}\\
  \midrule
  \tabincell{l}{$RI_{223}=-\frac{1}{6}RI_{13}(2RI_{44}+RI_{20}-RI^{2}_{4})$\\$~~~~~~~~~~~~+RI_{16}(RI_{44}-RI^{2}_{4}+RI_{20})-RI_{4}RI_{89}$\\$~~~~~~~~~~~~+RI_{165}+2RI_{208}-\frac{4}{3}RI_{209}$} & $RI_{224}=-RI_{4}(RI_{48}+RI_{106})$\\
  \midrule
  $RI_{225}=RI^{2}_{4}RI_{25}$ & $RI_{227}=RI_{20}RI_{30}$\\
  \bottomrule
\end{tabular}
\label{table:4}
\end{table*}

\begin{table*}
\newcommand{\tabincell}[2]{\begin{tabular}{@{}#1@{}}#2\end{tabular}}
\renewcommand
\arraystretch{1.25}
\caption{The explicit expressions of thirty-four rotation differential invariants in the set $S_{P}(3, 4)$.}
\centering
\begin{tabular}{ll}
  \toprule
  \textbf{Expressions} & \textbf{Expressions}\\
  \midrule
  $RI_{1}=f_{20}+f_{02}$&
  $RI_{2}=f^{2}_{10}+f^{2}_{01}$\\
  \midrule
  $RI_{3}=2f_{02}f_{20}-2f^{2}_{11}$&
  $RI_{4}=f_{40}+2f_{22}+f_{04}$\\
  \midrule
  $RI_{6}=f^{2}_{01}f_{02}+2f_{01}f_{10}f_{11}+f^{2}_{10}f_{20}$&
  $RI_{9}=f_{01}f_{03}+f_{01}f_{21}+f_{10}f_{12}+f_{10}f_{30}$\\
  \midrule
  $RI_{10}=f_{01}f_{12}+f_{01}f_{30}-f_{10}f_{03}-f_{10}f_{21}$&
  $RI_{11}=f^{2}_{01}f_{11}-f_{01}f_{02}f_{10}+f_{01}f_{10}f_{20}-f^{2}_{10}f_{11}$\\
  \midrule
  $RI_{13}=f^{2}_{03}+3f^{2}_{12}+3f^{2}_{21}+f^{2}_{30}$&
  $RI_{16}=2f_{03}f_{21}-2f^{2}_{12}+2f_{12}f_{30}-2f^{2}_{21}$\\
  \midrule
  $RI_{20}=f^{2}_{04}+4f^{2}_{13}+6f^{2}_{22}+4f^{2}_{31}+f^{2}_{40}$&
  \tabincell{l}{$RI_{21}=4f_{04}f_{22}f_{40}-4f_{04}f^{2}_{31}-4f^{2}_{13}f_{40}$\\$~~~~~~~~~~+8f_{13}f_{22}f_{31}-4f^{3}_{22}$}\\
  \midrule
  \tabincell{l}{$RI_{25}=f_{02}f_{04}+f_{02}f_{22}+2f_{11}f_{13}+2f_{11}f_{31}+f_{20}f_{22}$\\$~~~~~~~~~~+f_{20}f_{40}$}&
  \tabincell{l}{$RI_{26}=f^{2}_{01}f_{04}+f^{2}_{01}f_{22}+2f_{01}f_{10}f_{13}+2f_{01}f_{10}f_{31}+f^{2}_{10}f_{22}$\\$~~~~~~~~~~+f^{2}_{10}f_{40}$}\\
  \midrule
  \tabincell{l}{$RI_{29}=f_{01}f_{02}f_{03}+2f_{01}f_{11}f_{12}+f_{01}f_{20}f_{21}$\\$~~~~~~~~~~+f_{02}f_{10}f_{12}+2f_{10}f_{11}f_{21}+f_{10}f_{20}f_{30}$}&   \tabincell{l}{$RI_{30}=f_{02}f_{13}+f_{02}f_{31}-f_{04}f_{11}+f_{40}f_{11}-f_{13}f_{20}$\\~~~~~~~~~~$-f_{20}f_{31}$}\\
  \midrule
  \tabincell{l}{$RI_{31}=f^{2}_{01}f_{13}+f^{2}_{01}f_{31}-f_{01}f_{04}f_{10}+f_{01}f_{10}f_{40}$\\$~~~~~~~~~~-f^{2}_{10}f_{13}-f^{2}_{10}f_{31}$}&
  \tabincell{l}{$RI_{33}=f_{01}f_{02}f_{12}+2f_{01}f_{11}f_{21}+f_{01}f_{20}f_{30}$\\$~~~~~~~~~~-f_{02}f_{03}f_{10}-2f_{10}f_{11}f_{12}-f_{10}f_{20}f_{21}$}\\
  \midrule
  \tabincell{l}{$RI_{39}=f^{2}_{02}f_{04}+4f_{02}f_{11}f_{13}+2f_{02}f_{20}f_{22}+4f^{2}_{11}f_{22}$\\$~~~~~~~~~~+4f_{11}f_{20}f_{31}+f^{2}_{20}f_{40}$}&
  \tabincell{l}{$RI_{44}=2f_{04}f_{22}+2f_{04}f_{40}-2f^{2}_{13}-4f_{13}f_{31}+2f^{2}_{22}+2f_{22}f_{40}$\\$~~~~~~~~~~-2f^{2}_{31}$}\\
  \midrule
  \tabincell{l}{$RI_{45}=f_{01}f_{02}f_{03}+f_{01}f_{02}f_{21}+f_{01}f_{11}f_{12}$\\$~~~~~~~~~~+f_{01}f_{11}f_{30}+f_{03}f_{10}f_{11}+f_{10}f_{11}f_{21}$\\$~~~~~~~~~~+f_{10}f_{12}f_{20}+f_{10}f_{20}f_{30}$}& \tabincell{l}{$RI_{46}=f_{01}f_{03}f_{11}+f_{01}f_{11}f_{21}+f_{01}f_{12}f_{20}+f_{01}f_{20}f_{30}$\\$~~~~~~~~~~-f_{02}f_{03}f_{10}-f_{02}f_{10}f_{21}-f_{10}f_{11}f_{12}-f_{10}f_{11}f_{30}$}\\
  \midrule
  \tabincell{l}{$RI_{48}=f_{02}f_{03}f_{30}-f_{02}f_{12}f_{21}-2f_{03}f_{11}f_{21}$\\$~~~~~~~~~~-f_{03}f_{20}f_{30}+2f_{11}f^{2}_{12}+2f_{11}f_{12}f_{30}$\\$~~~~~~~~~~-2f_{11}f^{2}_{21}+f_{12}f_{20}f_{21}$}& \tabincell{l}{$RI_{54}=f_{02}f^{2}_{03}+2f_{02}f^{2}_{12}+f_{02}f^{2}_{21}+2f_{03}f_{11}f_{12}$\\$~~~~~~~~~~+4f_{11}f_{12}f_{21}+2f_{11}f_{21}f_{30}+f^{2}_{12}f_{20}+2f_{20}f^{2}_{21}$\\$~~~~~~~~~~+f_{20}f^{2}_{30}$}\\
  \midrule
  \tabincell{l}{$RI_{62}=-f_{02}f_{03}f_{21}+f_{02}f^{2}_{12}+f_{02}f_{12}f_{30}-f_{02}f^{2}_{21}$\\$~~~~~~~~~~-2f_{03}f_{11}f_{30}+f_{03}f_{20}f_{21}+2f_{11}f_{12}f_{21}$\\$~~~~~~~~~~-f^{2}_{12}f_{20}-f_{12}f_{20}f_{30}+f_{20}f^{2}_{21}$}& \tabincell{l}{$RI_{78}=f_{01}f_{04}f_{12}+f_{01}f_{12}f_{22}+2f_{01}f_{13}f_{21}+2f_{01}f_{21}f_{31}$\\$~~~~~~~~~~+f_{01}f_{22}f_{30}+f_{01}f_{30}f_{40}-f_{03}f_{04}f_{10}-f_{03}f_{10}f_{22}$\\$~~~~~~~~~~-2f_{10}f_{12}f_{13}-2f_{10}f_{12}f_{31}-f_{10}f_{21}f_{22}-f_{10}f_{21}f_{40}$}\\
  \midrule \tabincell{l}{$RI_{79}=f_{01}f_{03}f_{13}+f_{01}f_{03}f_{31}-f_{01}f_{04}f_{12}+f_{01}f_{12}f_{40}$\\$~~~~~~~~~~-f_{01}f_{13}f_{21}-f_{01}f_{21}f_{31}-f_{04}f_{10}f_{21}$\\$~~~~~~~~~~+f_{10}f_{12}f_{13}+f_{10}f_{12}f_{31}-f_{10}f_{13}f_{30}$\\$~~~~~~~~~~+f_{10}f_{21}f_{40}-f_{10}f_{30}f_{31}$}& \tabincell{l}{$RI_{83}=f_{01}f_{03}f_{22}+f_{01}f_{03}f_{40}+f_{01}f_{04}f_{21}-2f_{01}f_{12}f_{13}$\\$~~~~~~~~~~-2f_{01}f_{12}f_{31}+f_{01}f_{21}f_{22}+f_{04}f_{10}f_{30}+f_{10}f_{12}f_{22}$\\$~~~~~~~~~~+f_{10}f_{12}f_{40}-2f_{10}f_{13}f_{21}-2f_{10}f_{21}f_{31}+f_{10}f_{22}f_{30}$}\\
  \midrule \tabincell{l}{$RI_{89}=f^{2}_{03}f_{04}+4f_{03}f_{12}f_{13}+2f_{03}f_{21}f_{22}+f_{04}f^{2}_{12}$\\$~~~~~~~~~~+4f^{2}_{12}f_{22}+4f_{12}f_{13}f_{21}+4f_{12}f_{21}f_{31}$\\$~~~~~~~~~~+2f_{12}f_{22}f_{30}+4f^{2}_{21}f_{22}+f^{2}_{21}f_{40}+4f_{21}f_{30}f_{31}$\\$~~~~~~~~~~+f^{2}_{30}f_{40}$}& \tabincell{l}{$RI_{92}=-f_{02}f_{04}f_{22}+f_{02}f^{2}_{13}+f_{02}f_{22}f_{40}-f_{02}f^{2}_{31}$\\$~~~~~~~~~~-2f_{04}f_{11}f_{31}+f_{04}f_{20}f_{22}+2f_{11}f_{13}f_{22}-2f_{11}f_{13}f_{40}$\\$~~~~~~~~~~+2f_{11}f_{22}f_{31}-f^{2}_{13}f_{20}-f_{20}f_{22}f_{40}+f_{20}f^{2}_{31}$}\\
  \midrule \tabincell{l}{$RI_{95}=-f_{03}f_{04}f_{21}-2f_{03}f_{13}f_{30}+f_{03}f_{21}f_{40}$\\$~~~~~~~~~~-2f_{03}f_{30}f_{31}+f_{04}f^{2}_{12}+f_{04}f_{12}f_{30}-f_{04}f^{2}_{21}$\\$~~~~~~~~~~-f^{2}_{12}f_{40}+2f_{12}f_{13}f_{21}+2f_{12}f_{21}f_{31}$\\$~~~~~~~~~~-f_{12}f_{30}f_{40}+f^{2}_{21}f_{40}$}& \tabincell{l}{$RI_{97}=-f^{2}_{04}f_{31}+3f_{04}f_{13}f_{22}-f_{04}f_{13}f_{40}+3f_{04}f_{22}f_{31}$\\$~~~~~~~~~~+f_{04}f_{31}f_{40}-2f^{3}_{13}-2f^{2}_{13}f_{31}-3f_{13}f_{22}f_{40}+2f_{13}f^{2}_{31}$\\$~~~~~~~~~~+f_{13}f^{2}_{40}-3f_{22}f_{31}f_{40}+2f^{3}_{31}$}\\
  \midrule \tabincell{l}{$RI_{106}=-f_{02}f_{03}f_{12}-f_{02}f_{03}f_{30}-f_{02}f_{12}f_{21}$\\$~~~~~~~~~~~-f_{02}f_{21}f_{30}+f^{2}_{03}f_{11}+2f_{03}f_{11}f_{21}+f_{03}f_{12}f_{20}$\\$~~~~~~~~~~~+f_{03}f_{20}f_{30}-f_{11}f^{2}_{12}-2f_{11}f_{12}f_{30}+f_{11}f^{2}_{21}$\\$~~~~~~~~~~~+-f_{11}f^{2}_{30}+f_{12}f_{20}f_{21}$}& \tabincell{l}{$RI_{151}=f^{2}_{03}f_{04}+f^{2}_{03}f_{22}+f_{03}f_{04}f_{21}+2f_{03}f_{12}f_{13}$\\$~~~~~~~~~~~+2f_{03}f_{12}f_{31}+2f_{03}f_{21}f_{22}+f_{03}f_{21}f_{40}+f_{04}f^{2}_{12}$\\$~~~~~~~~~~~+f_{04}f_{12}f_{30}+f^{2}_{12}f_{22}+4f_{12}f_{13}f_{21}+4f_{12}f_{21}f_{31}$\\$~~~~~~~~~~~+2f_{12}f_{22}f_{30}+f_{12}f_{30}f_{40}+2f_{13}f_{21}f_{30}+f^{2}_{21}f_{22}$\\$~~~~~~~~~~~+f^{2}_{21}f_{40}+2f_{21}f_{30}f_{31}+f_{22}f^{2}_{30}+f^{2}_{30}f_{40}$}\\
  \bottomrule
\end{tabular}
\label{table:5}
\end{table*}

\subsection{Functional Dependencies}
\label{sec:5.3}
In fact, linear dependencies and polynomial dependencies are two special cases of functional dependencies. Previous authors have proved the number of functionally independent invariants should equal the number of independent variables (partial derivatives $f_{pq}$ for differential invariants) minus the number of free parameters of the transformation group (which is $1$ for rotations of 2D spatial coordinates). For any invariants in $S_{P}(4, 4)$, we have $1\leq (p+q)\leq 4$. Thus, there are $(2+3+4+5)=14$ independent partial derivatives and at most $14-1=13$ invariants in this set are functionally independent.

Given a dependent set $S=\{I_{1},I_{2},...,I_{n}\}$, mathematicians have proposed a simple method \cite{3} to eliminate functional dependencies among $I_{1}\sim I_{n}$. First, there must exist function $H$ such that $H(I_{1},I_{2},...,I_{n})=0$. Suppose each of $I_{i}$ is a functions of multiple variables $x_{1},x_{2},...,x_{m}$. Then, the function $H$ must hold
\begin{equation}\label{equ:30}
\begin{split}
\frac{\partial{H}}{\partial{x_{j}}}=\sum^{n}_{i=1}\frac{\partial{H(I_{1},I_{2},...,I_{n})}}{\partial{I_{i}}}\frac{\partial{I_{i}}}{\partial{x_{j}}}=0
\end{split}
\end{equation}
where $j=1,2,...,m$. Since the explicit expression of $I_{i}$ is known, the factor $\frac{\partial{I_{i}}}{\partial{x_{j}}}$ can be derived. The factor $\frac{\partial{H(I_{1},I_{2},...,I_{n})}}{\partial{I_{i}}}$ is unknown, but it is the same for all $j$. Thus, (\ref{equ:30}) can be regarded as a system of linear equations with the matrix of elements $a_{ij}=\frac{\partial{I_{i}}}{\partial{x_{j}}}$ and the vector of unknown coefficients $b_{j}=\frac{\partial{H}}{\partial{x_{j}}}$. If $I_{1}\sim I_{n}$ are independent, this system only has one solution with $b_{j}=0$ for all $l$. This means that the rank $n_{r}$ of the matrix $(a_{ij})$ must be $n$. If $n_{r}$ is less than $n$, only $n_{r}$ elements in the set $S$ are functionally independent.

We implement this method by using symbolic computing softwares, and finally derive thirteen functionally independent invariants from the set $S_{P}(4, 4)$, namely $RI_{1}$, $RI_{2}$, $RI_{3}$, $RI_{4}$, $RI_{6}$, $RI_{9}$, $RI_{10}$, $RI_{13}$, $RI_{20}$, $RI_{21}$, $RI_{25}$, $RI_{29}$ and $RI_{30}$. The set of them is denoted as $S_{F}(4, 4)$.

\subsection{Twelve Independent Sets}
\label{sec:5.4}
Given an independent set of rotation differential invariants, any subsets of this set have the same type of independence. For example, all invariants up to the order $3$ in the linearly independent set $S_{L}(4, 4)$ form a new linearly independent set, namely $S_{L}(4, 3)$. Similarly, the set of invariants up to the degree $3$ can be denoted as $S_{L}(3, 4)$. As an instance, the explicit expressions of invariants in the polynomially independent set $S_{P}(3,4)$ are shown in Table.~\ref{table:5}. In this way, we finally derive twelve independent sets, and the number of invariants in each one is listed in Table \ref{table:6}. Particularly, we find that
\begin{equation}\label{equ:31}
\begin{split}
&S_{F}(4, 4)=S_{F}(3, 4),~~S_{F}(4, 3)=S_{F}(3, 3)\\
\end{split}
\end{equation}

\begin{table}
\caption{The number of rotation differential invariants in twelve independent sets.}
\centering
\begin{tabular}{p{1.3cm}p{2.0cm}p{1.3cm}p{2.0cm}}
  \toprule
  \textbf{The set} & \textbf{The number of invariants} & \textbf{The set} & \textbf{The number of invariants}\\
  \midrule
  $S_{L}(4,4)$&
  $230$&
  $S_{L}(4,3)$&
  $64$\\
  $S_{L}(3,4)$&
  $59$&
  $S_{L}(3,3)$&
  $25$\\
  \midrule
  $S_{P}(4,4)$&
  $96$&
  $S_{P}(4,3)$&
  $30$\\
  $S_{P}(3,4)$&
  $34$&
  $S_{P}(3,3)$&
  $17$\\
  \midrule
  $S_{F}(4,4)$&
  $13$&
  $S_{F}(4,3)$&
  $8$\\
  $S_{F}(3,4)$&
  $13$&
  $S_{F}(3,3)$&
  $8$\\
  \bottomrule
\end{tabular}
\label{table:6}
\end{table}

\section{Experiment and Discussions}
\label{sec:6}
In this section, we first conducted patch verification on HPatches database. Twelve independent sets of rotation differential invariants generated in Sect.\ref{sec:5.4} were used as feature vectors. We analysed the influence of various factors on the performance of these invariant vectors. These factors include different types of independence among invariants, and their orders and degrees. Then, texture classification were performed on CUReT databases. We compared the performance of our homogeneous invariants with the corresponding non-polynomial invariants proposed in previous papers. In two experiments, some commonly used image features were also chosen for benchmark.

\subsection{Patch Verification}
\label{sec:6.1}
\begin{figure}
  \centering
  \includegraphics[height=60mm,width=60mm]{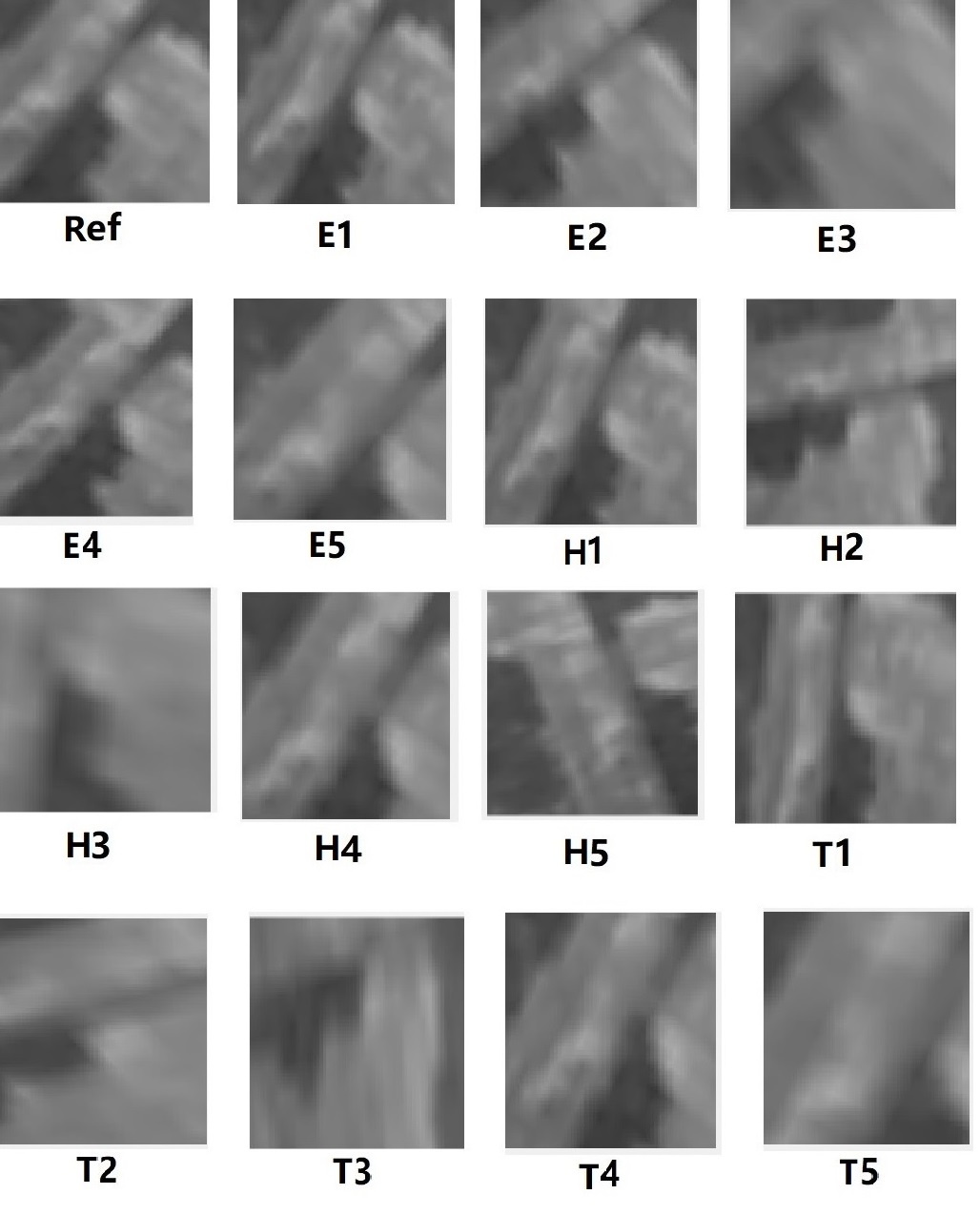}
\caption{One reference patch and fifteen target patches.}
\label{fig:1}
\end{figure}

The HPatches database \cite{2} consists of $65\times 65$ image patches, which are cropped from varying image sequences and are organized into pairs. An image sequence includes one reference image and five target images with geometric deformations and intensity variations. For patch verification, image features are employed to classify whether two patches in correspondence (positive pairs) or not (negative pairs). Note that negative pairs can be formed by patches from the same image sequence (Intra) and from different image sequences (Inter). It is obvious that the ones from the same image sequence are considered more challenging as they have similar textures. In order to increase the difficulty of this task, these patches are further disturbed by random generated easy (E), hard (H) and tough (T) image transformations. As an example, one reference patch and fifteen target patches in this database are shown in Fig.~\ref{fig:1}. We can find that most of target patches can be regarded as rotated versions of the reference patch.

\begin{figure}
  \centering
  \includegraphics[height=50mm,width=60mm]{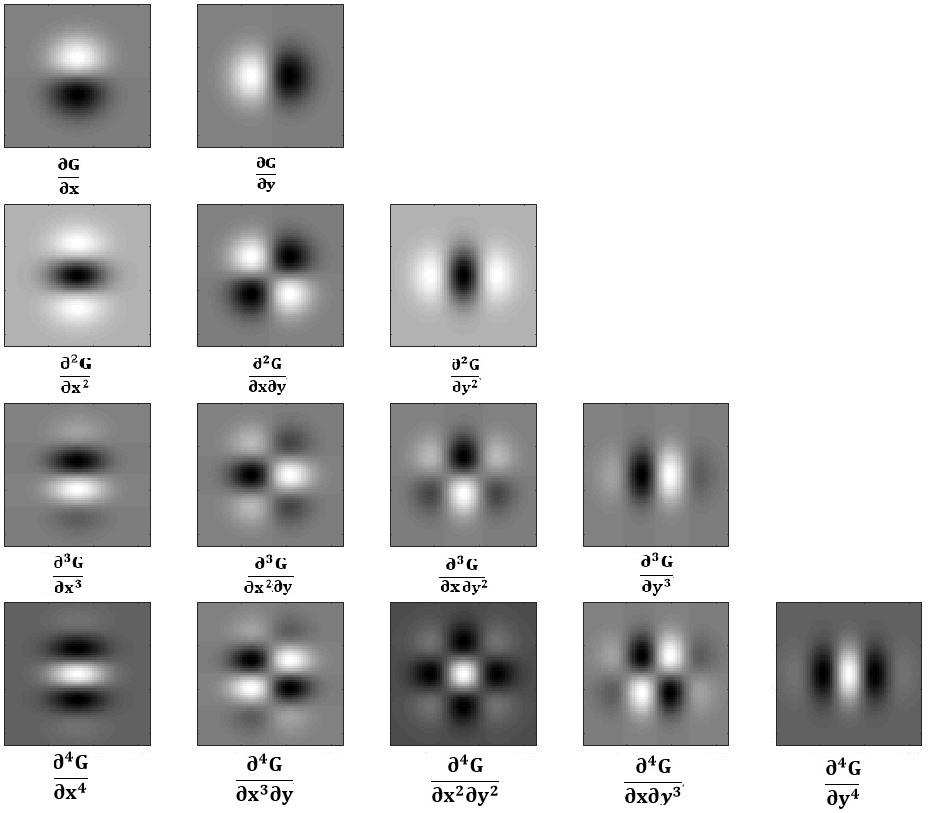}
\caption{The $65\times 65$ images of partial derivatives of Gaussian function $G(x,y;12)$ up to the order $4$.}
\label{fig:2}
\end{figure}

Twelve independent sets of rotation differential invariants generated in Section.\ref{sec:5.4} were used as feature vectors of each patch. We standardized all image patches to zero mean and unit variance before calculating invariants on them. This pre-processing operation can eliminate the effect of intensity transformations. In Section \ref{sec:3} and \ref{sec:4}, we suppose that image $f(x,y)$ is a continuous function and infinite-order partial derivatives exists. However, in practice, image $f(x,y)$ is a discrete function, making it impossible to directly obtain the values of those partial derivatives. Therefore, certain procedures are needed to estimate them. To solve this issue, we chose the Gaussian derivative method to estimate the values of partial derivatives of discrete image patches. The method uses partial derivatives of 2D Gaussian function $G(x,y;\sigma)$ as filters to form a convolution with the $65\times 65$ neighborhood of the central point $(33,33)$ on each patch. Note that, in this experiment, we set the scale factor $\sigma\in\{2,4,...,20\}$. The $65\times 65$ images of partial derivatives of Gaussian function $G(x,y;12)$ up to the order $4$ are shown in Fig.~\ref{fig:2}.

We followed the standard evaluation protocol provided by the authors and only replaced the Euclidean distance with the Chi-Square distance to measure the similarity between two image patches. In fact, the magnitudes of rotation differential invariants $RI_{1}\sim RI_{230}$ are not the same. If these invariants are not normalized, the distances between two feature vectors could be dominated by some of invariants instead of all of invariants. The Chi-Square distance can solve this problem naturally.

\begin{figure*}
  \centering
  \subfloat[Four linearly independent sets.]{\includegraphics[height=85mm,width=140mm]{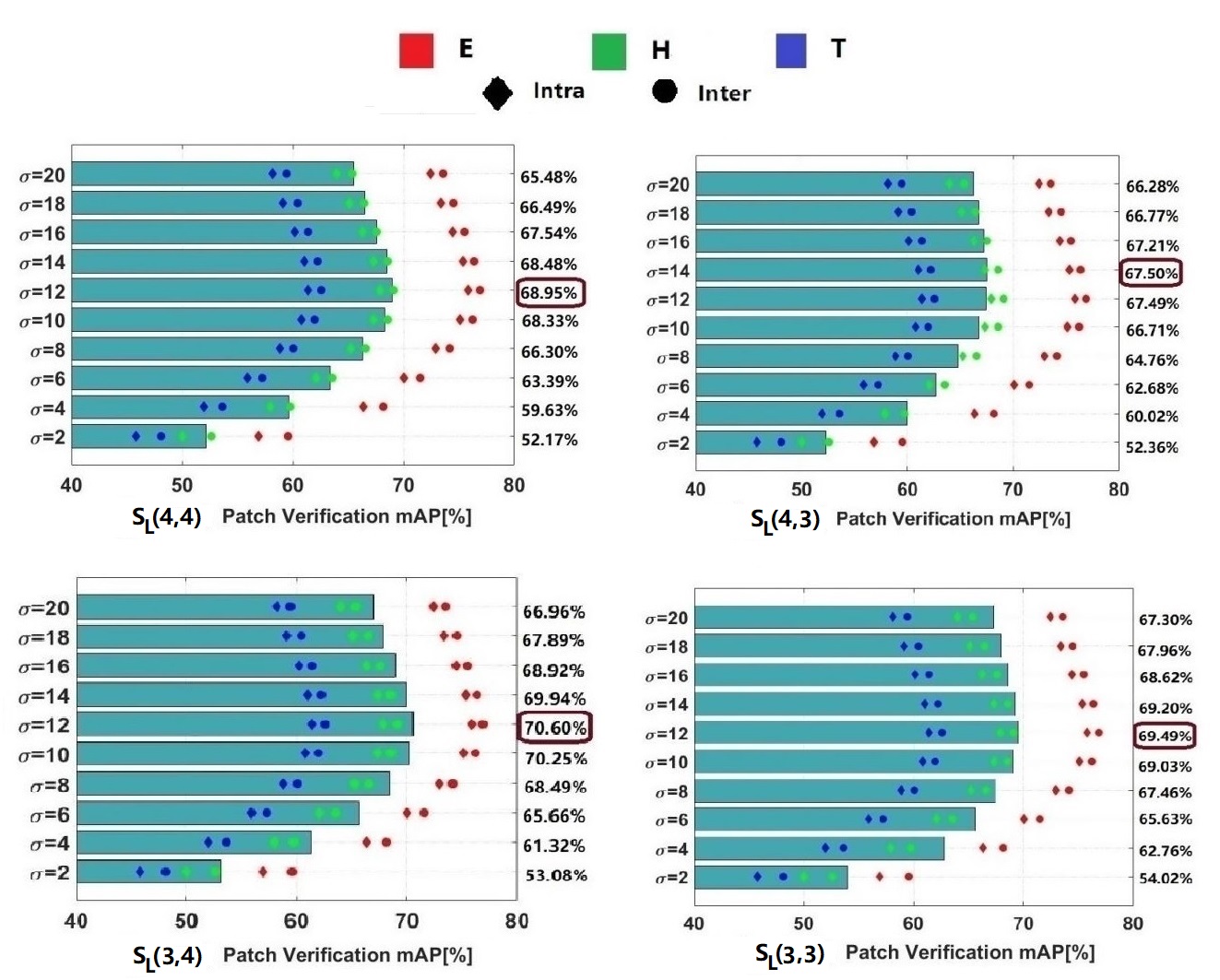}\label{figure:3(a)}\hfill}\\
  \subfloat[Four polynomially independent sets.]{\includegraphics[height=80mm,width=140mm]{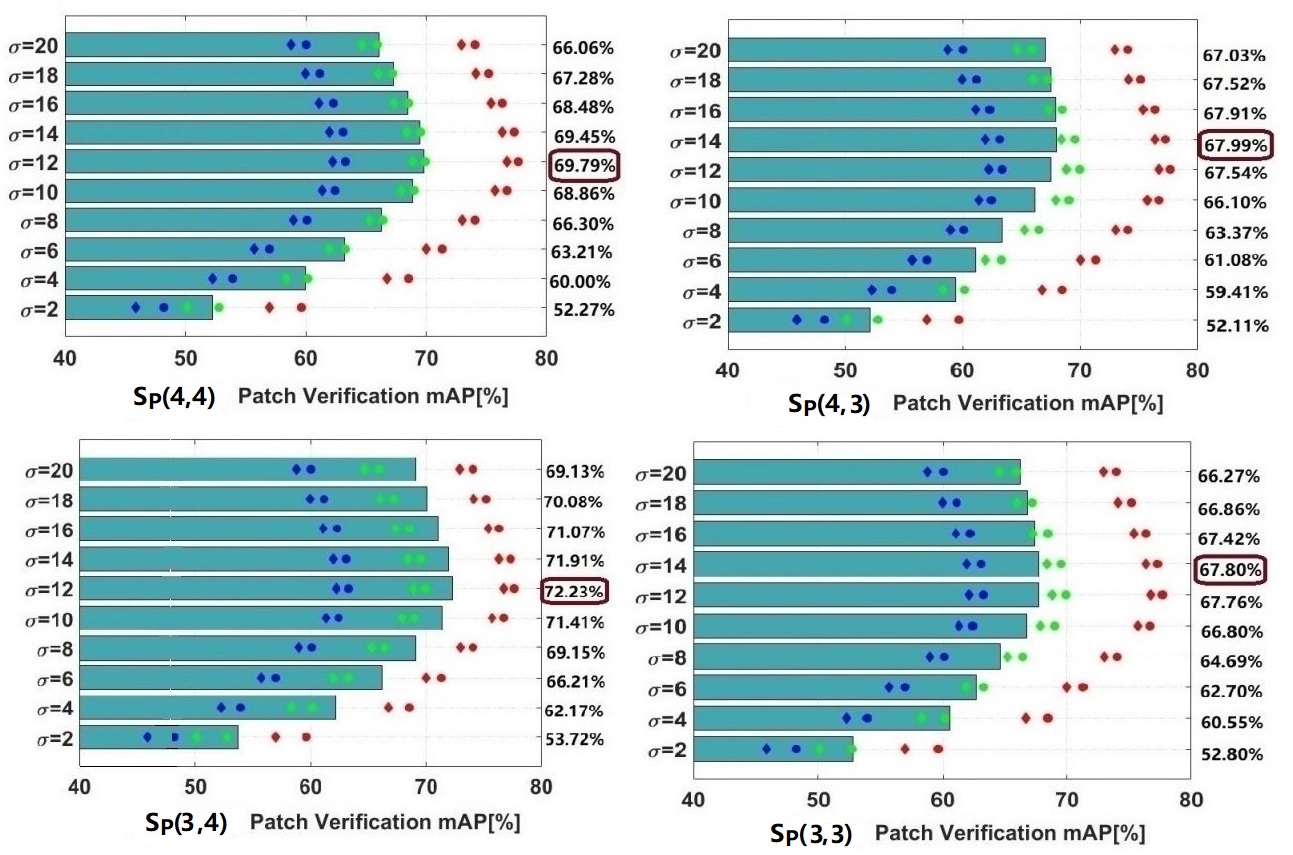}\label{figure:3(b)}\hfill}\\
  \subfloat[Two functionally independent sets.]{\includegraphics[height=35mm,width=140mm]{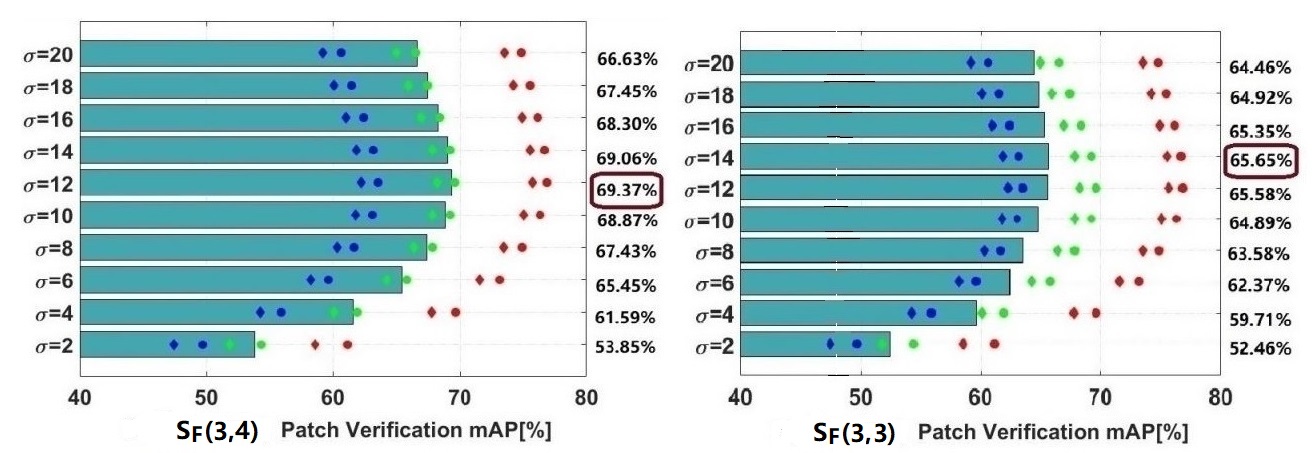}\label{figure:3(c)}\hfill}
  \caption{The mAP from twelve independent sets of rotation differential invariants on the HPatches database.}
  \label{fig:3}
\end{figure*}

The mean average precision(mAP) is employed to evaluate the performance of twelve invariant vectors. The results are shown in Fig.~\ref{fig:3}. Since $S_{F}(4,4)$=$S_{F}(3,4)$ and $S_{F}(4,3)$=$S_{F}(3,3)$, the mAP from them were omitted. There are several interesting consequences:
\begin{itemize}
  \item The accuracy from each invariant vector first increased and then declined as the parameter $\sigma$, namely the scale factor of 2D Gaussian function, gradually changed from $2$ to $20$. In most cases, the highest accuracy was acquired by setting $\sigma=12$. As stated previously, the size of partial derivatives of 2D Gaussian function is also $65\times 65$. According to $3\sigma$ rule of thumb, we should set $\sigma=\frac{65}{6}\approx 11$ to ensure that nearly $100\%$ value of these derivatives lies within the $65\times65$ region.
  \item The performance of $S_{P}(N,K)$ is better than the corresponding $S_{L}(N,K)$ and $S_{F}(N,K)$, where $N,K\in\{3,4\}$. First, the number of invariants in $S_{F}(N,K)$ is too few to provide enough discrimination power. Then, although he number of invariants in $S_{L}(N,K)$ is more than the corresponding $S_{P}(N,K)$, those invariants which can be expressed as polynomials of the others not only do not extract new information of image patches, but increase computational errors.
  \item The accuracy from $S_{A}(N,4)$ is higher than that from $S_{A}(N,3)$, where the subscript $A\in\{L,P,F\}$ and $N\in\{3,4\}$. This is due to high-order partial derivatives extracting more information of image local structures. On the contrary, $S_{A}(3,K)$ performs better than $S_{A}(4,K)$, where $K\in\{3,4\}$. In fact, previous researches have claimed that the numerical stability of invariants deteriorates rapidly as their orders increase.
\end{itemize}

\begin{figure}
  \centering
  \includegraphics[height=35mm,width=85mm]{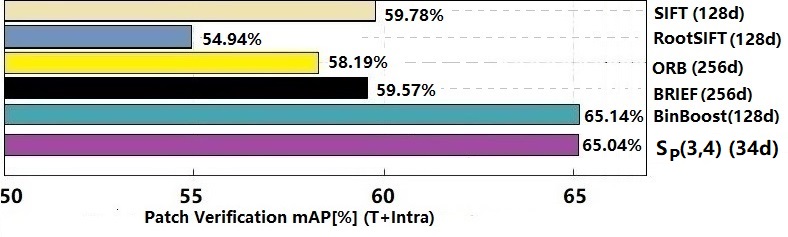}
\caption{The mAP from various image features on (T+Intra) subdatabases.}
\label{fig:4}
\end{figure}

Five famous handcrafted image features were also chosen for benchmark, namely SIFT \cite{6}, RootSIFT \cite{1}, ORB \cite{22}, BRIEF \cite{4} and BinBoost \cite{25}. In order to increase the difficultly of the experiment, patch verification was carried out on (T+Intra) subdatabases. The mAP from them and $S_{P}(3,4)$ is displayed in Fig.~\ref{fig:4}. Clearly, when image patches are deformed by tough image transformations, the performance of $S_{P}(3,4)$ is superior to almost all of image features. The only feature that outperforms $S_{P}(3,4)$ is BinBoost. However, it is constructed by using a complicated learning framework. In addition, we should note that the dimension of BinBoost is 7 times the dimension of $S_{P}(3,4)$. This implies that the computational cost of $S_{P}(3,4)$ is lower than that of BinBoost.

\subsection{Texture Classification}
\label{sec:6.2}
\begin{figure}
  \centering
  \includegraphics[height=30mm,width=80mm]{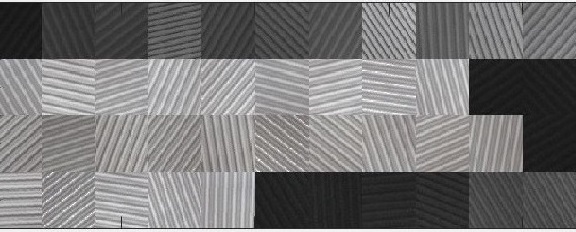}
\caption{Some texture classes in the rot-CUReT database.}
\label{fig:5}
\end{figure}

This experiment was designed to test the performance of homogeneous invariants and the corresponding non-polynomial invariants on the classification of texture images. We used the CUReT database available at \url{www.robots.ox.ac.uk/~vgg/research/texclass}. It contains 61 texture classes with 92 images per class, and the size of each image is $200\times 200$. In fact, this database is not very suitable to evaluate the invariance of image features to rotation, because most of images in the same class have no significant rotation. To resolve this problem, we first randomly rotated each image in the database. Then, forth-six training images are randomly selected from each of 61 classes, and the rest images were testing images. Some images from the same class are shown in Fig.~\ref{fig:5}.

In order to eliminate the influence of intensity transformations, we also use the pre-processing operation to normalize each of images in this database. And then, seven texture features were calculated. Three of them are constructed by using the same rotation differential invariants.

\textbf{RI(12d)}: We use the first three homogeneous invariants shown in Table.~\ref{table:5}, namely $RI_{1}$, $RI_{2}$ and $RI_{3}$. Similar to Section \ref{sec:6.1}, the Gaussian derivative method was used to produce the numerical values of partial derivatives at each point on a texture image. Four 3-dimensional invariant vectors at each point can be derived by setting the scale factor $\sigma=1,2,4,8$, respectively. They are combined to construct a 12-dimensional texture feature.

\textbf{BIF (1296d)}: BIF \cite{10} consists of six functions of homogeneous invariants $RI_{1}$, $RI_{2}$ and $RI_{3}$: $2\sqrt{RI_{2}}$, $\pm RI_{1}$, $\frac{1}{\sqrt{2}}\left(\sqrt{RI^{1}_{2}-2RI_{3}}\pm RI_{1}\right)$, $\sqrt{RI^{1}_{2}-2RI_{3}}$. They can be regarded as six non-polynomial invariants. Using the derivatives of 2D Gaussian function $G(x,y;\sigma)$ as filters, the value of BIF was calculated at each point on an image. If the value of the $i$th function was larger than the others, the point was labelled as $i$, where $i\in \{1,2,...,6\}$. Griffin et~al. set the scale factor $\sigma=1,2,4,8$, and obtained a 4-dimensional label at each point. This label has a total of $6^{4}=1296$ possible values. Thus, they finally derived a 1296-dimensional histogram of labels.

\textbf{CMR (8d)}: Zhang et~al. \cite{28} calculate the maximum values of the first and the second directional derivatives at each point on an image. We find that two maximum values are also non-polynomial invariants: $RI_{2}$, $\frac{1}{\sqrt{2}}\left(RI_{1}+\sqrt{RI^{2}_{1}-2RI_{3}}\right)$. They were calculated by setting $\sigma=1,2,4,8$, respectively, and then combined to a 8-dimensional descriptor.

\textbf{LBP$^{riu2}_{24,3}$ (26d)}: The rotation-invariant uniform local binary patterns (24 sampling neighbors on a circle of radius 3) were proposed in \cite{19}. It is the most commonly used texture feature. Given an image, a 26-dimensional histogram of binary patterns was obtained.

\textbf{VZ-MR8 (8d)}: This method \cite{26} extracted image information in the neighborhood of each point by using the MR8 filter bank and generated a 8-dimensional image feature at each point. This feature is only invariant to six special rotations, namely the angle of rotation $\theta=\frac{2\pi}{6}\cdot k$, where $k=1,2,...,6$.

\textbf{VZ-Joint (49d)}: The image patch in the $7\times 7$ square neighborhood of each point was used to describe image local structures \cite{27}. Note that this 49-dimensional feature is not invariant to rotation.

\textbf{S-C (13d)}: The responses of thirteen filters which have circular symmetry were used to construct a 13-dimensional feature at each point \cite{23}.

We derived the values of RI, CMR, VZ-MR8, VZ-Joint and S-C at each point on a texture image, and used the procedure published in \cite{26} to construct the corresponding histograms to describe overall information of the image. Specifically, thirteen images were chosen randomly for each texture. The image features calculated at each point on each image were clustered by k-means clustering, and produced 10 texton cluster centres. Finally, the texton dictionary were generated, which contains $10\cdot 61=610$ textons. We made use of it to label each point on a given image and derived a 610-dimensional histogram of texton frequencies.

The Nearest Neighbor classifier is used for classification. We repeated the classification experiment $100$ times by randomly selected training images. The mean accuracy rates from various features is shown in Table.~\ref{table:7}. First, we can find that the performance of RI, BIF and CMR is better than the others. This shows that rotation differential invariants are more suitable to describe texture images, because they can capture intrinsic information of image local structures. As mentioned above, both BIF and CMR can be expressed as the functions of three homogenous invariants used in RI. An interesting observation is that the classification accuracy from RI ($97.14\pm0.80\%$) is higher than that from BIF ($96.69\pm0.62\%$) and CMR ($96.84\pm0.77\%$). This seems to indicate that we do not have to construct the complicated functions of homogeneous invariants when we use the texton-based approach to construct image features. In fact, homogeneous invariants produced in our paper are more fundamental than non-polynomial invariants published in previous papers.

\begin{table}
\renewcommand
\arraystretch{1.5}
\caption{The classification accuracy rates from various image features on the CUReT database.}
\centering
\begin{tabular}{p{30mm}p{30mm}}
  \toprule
  \textbf{Image Features} & \textbf{Accuracy Rates}\\
  \midrule
  LBP$^{riu2}_{24,3}$  (26d) & $71.88\pm1.13\%$\\
  VZ-Joint (610d)& $72.48\pm0.98\%$\\
  S-C (610d)& $93.71\pm1.19\%$\\
  VZ-MR8 (610d)& $95.67\pm0.94\%$\\
  \midrule
  BIF (1296d) & $96.69\pm0.62\%$\\
  CMR (610d)& $96.84 \pm 0.77\%$\\
  RI (610d)& \textbf{97.14}$\pm$\textbf{0.80}$\%$\\
  \bottomrule
\end{tabular}
\label{table:7}
\end{table}

\section{Conclusion and Future Work}
\label{sec:7}
In this paper, we design two fundamental differential operators to automatically generate high-order rotation differential invariants of images, which can be expressed as homogeneous polynomials in image partial derivatives. We derive a series of explicit instances of homogeneous invariants and analysed the dependencies among them in detail. Image classification and verification were carried out on real image databases. Based on the experimental results, we analysed the influence of various factors on the performance of rotation differential invariants, and find these invariants performed better than commonly used image features in some cases.

In the future, we plan to analyse geometric meaning of high-order rotation differential invariants and construct more effective image features based on them. We also want to combine image differential invariants with convolutional neural networks to improve the accuracy of image classification or retrieval.

\begin{acknowledgements}
This work has partly been funded by the National Key R$\&$D Program of China (No. 2017YFB1002703) and the National Natural Science Foundation of China (Grant No. 60873164, 61227802 and 61379082).
\end{acknowledgements}


\begin{thebibliography}{32}
\bibitem{1}
Arandjelovi$\acute{c}$, R., Zisserman, A.: Three things everyone should know to improve object retrieval. In: Proceedings of the IEEE Conference on Computer Vision and Pattern Recognition, pp. 2911-2918 (2012)

\bibitem{2}
Balntas, V., Lenc, K., Vedaldi, A., Mikolajczyk, K: HPatches: A benchmark and evaluation of handcrafted and learned local descriptors. In: Proceedings of the IEEE Conference on Computer Vision and Pattern Recognition, pp. 5173-5182 (2017)

\bibitem{3}
Browm, A.,B.: Functional dependence. Transactions of American Mathematical Society. \textbf{38}(2), 379-394 (1935)

\bibitem{4}
Calonder, M., Lepetit, V., Ozuysal, M., Trzcinski, T., Strecha, C., Fua, P.: Brief: computing a local binary descriptor very fast. IEEE Transactions on Pattern Analysis and Machine Intelligence, \textbf{34}(7), 1281-1298 (2012)

\bibitem{5}
Crosier, M., Griffin, L. D.: Using basic image features for texture classification. International Journal of Computer Vision, \textbf{88}(3), 447-460 (2010)

\bibitem{6}
David, G. L.: Distinctive image features from scale-invariant keypoints. International Journal of Computer Vision, \textbf{60}(2), 99-110 (2004)

\bibitem{7}
Florack, L. M. J., Ter Haar Romeny, B. M., Koenderink, J. J., Viergever, M. A.: Scale and the differential structure of images. Image and Vision Computing, \textbf{10}(6), 376-388 (1992)

\bibitem{8}
Florack, L. M. J., Ter Haar Romeny, B. M., Koenderink, J. J., Viergever, M. A.: Cartesian differential invariants in scale-space. Journal of Mathematical Imaging and Vision, \textbf{3}(4), 327-348 (1993)

\bibitem{9}
Griffin, L. D. The second order local-image-structure solid. IEEE Transactions on Pattern Analysis and Machine Intelligence, \textbf{29}(8), 1355-1366 (2007).

\bibitem{10}
Griffin, L. D., Lillholm, M., Crosier, M., Van Sande, J.: Basic image features (BIFs) arising from approximate symmetry type. In: Proceedings of the Scale Space and Variational Methods in Computer Vision, pp. 343-355 (2009)

\bibitem{11}
Griffin, L. D., Lillholm, M.: Symmetry sensitivities of derivative-of-Gaussian filters. IEEE Transactions on Pattern Analysis and Machine Intelligence, \textbf{32}(6), 1072-1083, (2010)

\bibitem{12}
Koenderink, J. J., Van Doorn, A. J. Surface shape and curvature scales. Image And Vision Computing, \textbf{10}(8), 557-564 (1992)

\bibitem{13}
Lindeberg, T.: On scale selection for differential operators. In: Proceedings of the Scale Space in Computer Vision, pp. 317-348 (1994)

\bibitem{14}
Lindeberg, T., G$\mathring{a}$rding, J.: Shape-adapted smoothing in estimation of 3-D shape cues from affine deformations of local 2-D brightness structure. Image And Vision Computing, \textbf{15}(6), 415-434 (1997)

\bibitem{15}
Mikolajczyk, K., Schmid, C.: Indexing based on scale invariant interest points. In: Proceedings of the IEEE International Conference on Computer Vision, pp. 525-531 (2001).

\bibitem{16}
Mikolajczyk, K., Schmid, C.: Scale and affine invariant interest point detectors. International Journal of Computer Vision, \textbf{60}(1), 63-86 (2004)

\bibitem{17}
Mikolajczyk, K., Schmid, C.: A performance evaluation of local descriptors. IEEE Transactions on Pattern Analysis and Machine Intelligence, \textbf{27}(10), 1615-1630 (2005)

\bibitem{18}
Mikolajczyk, K., Tuytelaars, T., Schmid, C., Zisserman, A., Matas, J., Schaffalitzky, F., Kadir, T., Van Gool, L. A comparison of affine region detectors. International Journal of Computer Vision, \textbf{65}(1-2), 43-72 (2005).

\bibitem{19}
Ojala, T., Pietikainen, M., Maenpaa, T.: Multiresolution gray-scale and rotation invariant texture classification with local binary patterns. IEEE Transactions on Pattern Analysis and Machine Intelligence, \textbf{24}(7), 971-987 (2002)

\bibitem{20}
Olver, P. J.: Equivalence, invariants, and symmetry. Cambridge University Press (1995)

\bibitem{21}
Olver, P. J.: Classical Invariant Theory. Cambridge University Press (1999)

\bibitem{22}
Rublee, E., Rabaud, V., Konolige, K., Bradski, G.: ORB: an effcient alternative to SIFT or SURF. In: Proceedings of the IEEE International Conference on Computer Vision, pp. 2564-2571 (2011)

\bibitem{23}
Schmid, C.: Constructing models for content-based image retrieval. In: Proceedings of the IEEE International Conference on Computer Vision and Pattern Recognition, pp. 39-45 (2001)

\bibitem{24}
Ter Haar Romeny, B. M., Florack, L. M. J., Salden, A. H., Viergever, M. A.: Higher order differential structure of images. Image and Vision Computing, \textbf{12}(6), 317-325 (1994)

\bibitem{25}
Trzcinski, T., Christoudias, M., Lepetit, V.: Learning image descriptors with boosing. IEEE Transactions on Pattern Analysis and Machine Intelligence, \textbf{37}(3), 597-610 (2014)

\bibitem{26}
Varma, M., Zisserman, A.: A statistical approach to texture classification from single images. International Journal of Computer Vision, \textbf{62}(1-2), 61-81 (2005)

\bibitem{27}
Varma, M., Zisserman, A. A statistical approach to material classification using image patch exemplars. IEEE Transactions on Pattern Analysis and Machine Intelligence, \textbf{31}(11), 2032-2047 (2009)

\bibitem{28}
Zhang, J., Zhao, H., Liang, J. M.: Continuous rotation invariant local descriptors for texton dictionary-based texture classification. Computer Vision and Image Understanding, \textbf{117}(1), 56-75 (2013)

\end{thebibliography}

\end{document}